\documentclass{article}

\usepackage{microtype}
\usepackage{graphicx}
\usepackage{subfigure}
\usepackage{booktabs} % for professional tables
\usepackage{multirow} % we add!

\usepackage{hyperref}

\usepackage[accepted]{icml2025}

\usepackage{amsmath}
\usepackage{amssymb}
\usepackage{mathtools}
\usepackage{amsthm}

\usepackage[capitalize,noabbrev]{cleveref}

\theoremstyle{plain}
\newtheorem{theorem}{Theorem}[section]

\theoremstyle{definition}

\theoremstyle{remark}

\usepackage[textsize=tiny]{todonotes}

\icmltitlerunning{Wide \text{\&} Deep Learning for Node Classification}

\begin{document}

\twocolumn[
\icmltitle{Wide \text{\&} Deep Learning for Node Classification}



\icmlsetsymbol{equal}{*}

\begin{icmlauthorlist}
\icmlauthor{Yancheng Chen}{sais,sms}
\icmlauthor{Wenguo Yang}{sms}
\icmlauthor{Zhipeng Jiang}{sms}
\end{icmlauthorlist}

\icmlaffiliation{sais}{School of Advanced Interdisciplinary Sciences, University of Chinese Academy of Sciences}
\icmlaffiliation{sms}{School of Mathematical Sciences, University of Chinese Academy of Sciences}

\icmlcorrespondingauthor{Zhipeng Jiang}{jiangzhipeng@ucas.ac.cn}

\icmlkeywords{Graph Convolutional Networks, Node Classification}

\vskip 0.3in
]



\printAffiliationsAndNotice{}  % leave blank if no need to mention equal contribution

\begin{abstract}
Wide \text{\&} Deep, a simple yet effective learning architecture for recommendation systems developed by Google, has had a significant impact in both academia and industry due to its combination of the memorization ability of generalized linear models and the generalization ability of deep models.
Graph convolutional networks (GCNs) remain dominant in node classification tasks; however, recent studies have highlighted issues such as {\em heterophily} and {\em expressiveness}, which focus on graph structure while seemingly neglecting the potential role of node features.
In this paper, we propose a flexible framework GCNIII, which leverages the Wide \text{\&} Deep architecture and incorporates three techniques: {\em Intersect memory}, {\em Initial residual} and  {\em Identity mapping}.
We provide comprehensive empirical evidence showing that GCNIII can more effectively balance the trade-off between {\em over-fitting} and {\em over-generalization} on various semi- and full- supervised tasks. 
Additionally, we explore the use of large language models (LLMs) for node feature engineering to enhance the performance of GCNIII in cross-domain node classification tasks.
Our implementation is available at \url{https://github.com/CYCUCAS/GCNIII}.
\end{abstract}

\section{Introduction}
\label{sec:intro}

\begin{figure}[ht]
    \begin{center} 
    \centerline{\includegraphics[width=\columnwidth]{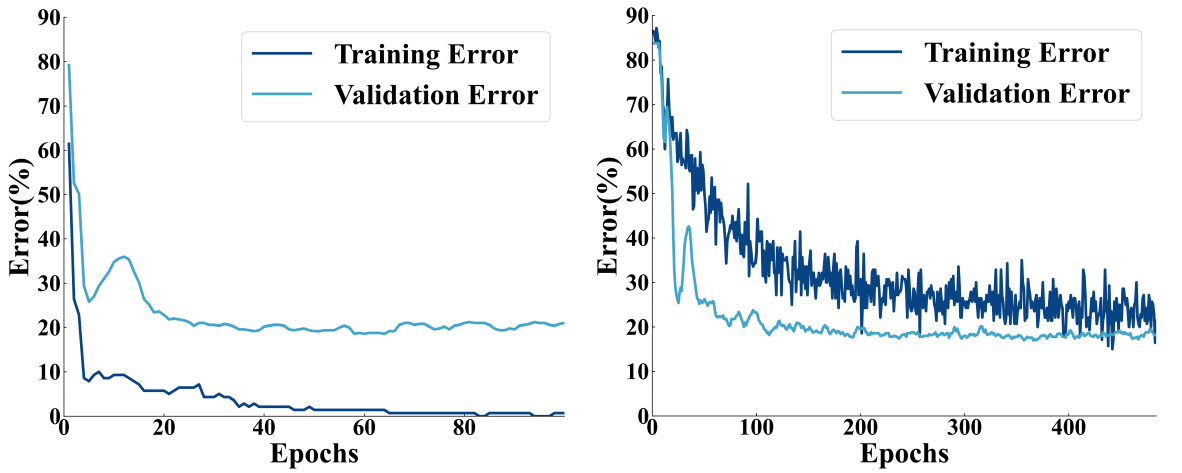}}
    \vspace{-3mm}
    \caption{Training error and validation error of the semi-supervised task on Cora with 2-layer vanilla GCN (left) and 64-layer GCNII (right). The training error of deep GCNII is very volatile and much higher than the validation error. We call this phenomenon {\em over-generalization}.}
    \label{train_val_error}
    \end{center}
    \vspace{-10mm}
\end{figure}

Node classification is a machine learning task in graph-structured data analysis~\cite{sen08}, where the goal is to assign labels to nodes in a graph based on their properties and the relationships between them.
While graph convolutional networks (GCNs)~\cite{kipf17} have achieved great success in node classification due to their strong generalization performance~\cite{xu21}, some studies have pointed out that message passing neural networks~\cite{gilmer17}, such as GCNs, have several limitations including {\em homophily} assumption~\cite{zhu20,luan22} and lack of {\em expressiveness}~\cite{xu19}.
However, recent studies~\cite{ma22, platonov23} have found that GCNs can also achieve strong results on heterophilous graphs.
Moreover, the latest work~\cite{luo24b} indicates that Graph Transformers (GTs)~\cite{ying21, ramp22, wu23, deng24}, which are theoretically proven to be more expressive~\cite{zhang23a,zhang23b}, do not outperform GCNs in node classification tasks.
In summary, GCNs remain dominant in node classification tasks.

The classic models, such as GCN~\cite{kipf17}, GAT~\cite{veli18} and GraphSAGE~\cite{hamilton17}, can achieve their best performance with 2-layer shallow models, and stacking more layers will significantly degrade performance.
There are at least two possible reasons for this phenomenon. 
One is {\em over-smoothing}~\cite{li18}, in which the embedding vector of the connected nodes becomes indistinguishable after multi-layer graph convolution; the other is that the parameters in the deep graph convolution layers are challenging to optimize~\cite{zhang21}.

\begin{figure}[ht]
    \begin{center}
    \centerline{\includegraphics[width=8cm]{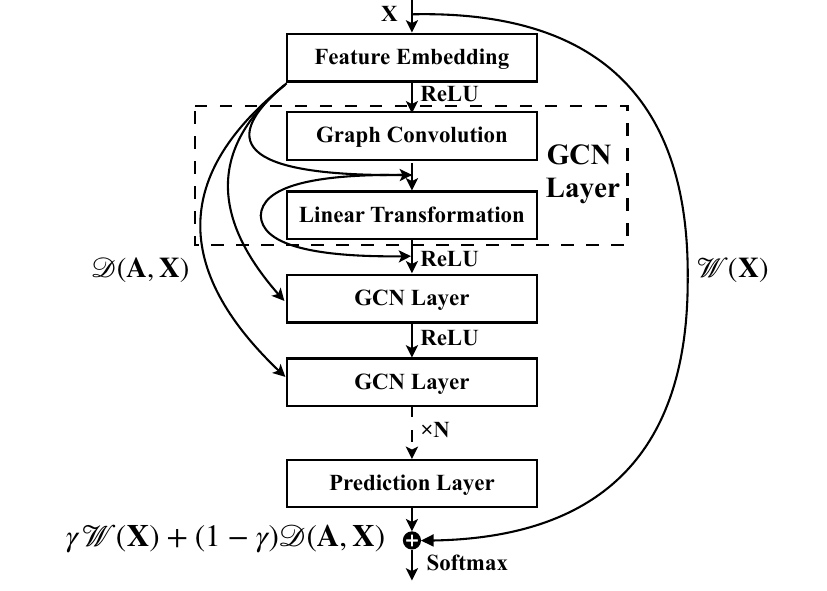}}
    \vspace{-3mm}
    \caption{Wide \text{\&} Deep architecture GCNIII.}
    \label{wide_and_deep}
    \end{center}
    \vspace{-10mm}
\end{figure}

Since shallow GCNs limit their ability to extract information from higher-order neighbors, many studies have explored ways to develop deeper models while relieving the problem of {\em over-smoothing}.
JK-Nets~\cite{xu18} use dense skip connections to flexibly leverage different neighborhood ranges.
SGC~\cite{wu19} removes nonlinearities and collapses weight matrices between consecutive layers by applying the $K$-th power of the graph convolution matrix in a single layer.
DropEdge~\cite{rong20} randomly removes a certain number of edges from the input graph at each training epoch, acting like a data augmenter and also a message passing reducer.
DAGNN~\cite{liu20} decoupling the entanglement of representation transformation and propagation in current graph convolution operations learns graph node representations by adaptively incorporating information from large receptive fields.

When \citet{kipf17} adapt {\em residual connection} ~\cite{he16} to GCN and PPNP~\cite{gasteiger19} uses a variant of personalized PageRank~\cite{page99} instead of graph convolution, GCNII~\cite{chen20} incorporates ideas from both and continues to achieve state-of-the-art performance to this day.
However, when we train 64-layer GCNII on a semi-supervised task, we find that the error on the validation set is much lower than training set as shown in \cref{train_val_error}, and this is quite different from the performance of 2-layer vanilla GCN which is easy to {\em over-fitting} on the training set.
This phenomenon is often referred to as {\em under-fitting}, but {\em under-fitting} models cannot perform well on validation and test sets. 
This is also not a phenomenon of {\em over-smoothing}, because {\em over-smoothing} is global, thus the error of the validation set cannot differ too much from the test set. 
Therefore, we call this curious phenomenon {\em over-generalization}, which has not been shown in any other study.

In conclusion, the role and mechanism of deep GCNs are not yet clear. 
When faced with different graph datas, it remains an open problem what type of GCN, whether shallow or deep, should be used.
Unlike many prior works that focused solely on graph structure, we also investigate the role of node features.
We propose \textbf{G}raph \textbf{C}onvolutional \textbf{N}etwork with \textbf{I}ntersect memory, \textbf{I}nitial residual and \textbf{I}dentity mapping (GCNIII), a Wide \text{\&} Deep architecture model as shown in \cref{wide_and_deep} that can more effectively balance the trade-off between {\em over-fitting} and {\em over-generalization} and achieves state-of-the-art performance on various semi- and full- supervised tasks.
\section{Preliminaries}
\label{sec:pre}

\paragraph{Node Classification.}
For node classification tasks, the input data is generally a simple and connected undirected graph \( \mathcal{G} =(\mathcal{V}, \mathcal{E} )\) with $n$ nodes. 
The information we can use for node classification includes structure information and feature information. 
Structure information is generally represented by adjacency matrix $\mathbf{A}$ and degree matrix $\mathbf{D}$, and the information of the latter is contained in the former.
Feature information is generally represented by node feature matrix $\mathbf{X} \in \mathbb{R}^{n \times d}$, which means that each node $v$ is associated with a $d$-dimensional row vector $\mathbf{x}_v$.
The goal is to build model $f$ such that the probability distributions $\mathcal{P}$ of the predicted node classes are as similar as possible to the real labels $\mathcal{C} $:
\begin{equation}
  \label{eqn:node_classification}
  \mathcal{P}  = f\left( \mathbf{A}, \mathbf{X}  \right).
\end{equation}
However, the two types of information in different datasets are different. 
Many previous studies, such as {\em heterophily} and {\em expressiveness}, focus on structural information, while in-depth studies on feature information are few.
For example, the most classic citation network datasets uses sparse features based on a bag-of-words representation of the document.
How does the sparsity or denseness of features affect the performance of node classification tasks? 
Our studies suggest that effectively leveraging both structure and feature information is the key to improving node classification performance.

\paragraph{Wide \text{\&} Deep.}
\citet{cheng16} suggest that {\em memorization} and {\em generalization} are both important for recommender systems. 
Wide linear models can effectively memorize sparse feature interactions using cross-product feature transformations, while deep neural networks can generalize to previously unseen feature interactions through low-dimensional embeddings.
They presented the Wide \text{\&} Deep learning framework to combine the strengths of both types of model.
For a logistic regression problem, the model’s prediction is:
\begin{equation}
  \label{eqn:widedeep}
  \Pr(\mathbf{y} = 1|\mathbf{x}) = \sigma\left( \mathbf{w}_{w}^{T} [\mathbf{x}, \phi(\mathbf{x}) ] + \mathbf{w}_{d}^{T}\mathbf{x}^{(n)} + b \right),
\end{equation}
where $\mathbf{y}$ is the binary class label, $\sigma(\cdot )$ is the sigmoid function, $\phi(\mathbf{x})$ are the cross product transformations of the original features $\mathbf{x}$, and $b$ is the bias term. 
$\mathbf{w}_{w}$ is the vector of all wide model weights, and $\mathbf{w}_{d}$ are the weights applied on the final embedding $\mathbf{x}^{(n)}$, which is obtained by the iteration $\mathbf{x}^{(l+1)} = ReLU(\mathbf{W}^{(l)}\mathbf{x}^{(l)} + b^{(l)})$ of a feed-forward neural network.

\paragraph{GCN.}
\citet{kipf17} propose a multi-layer Graph Convolutional Network (GCN) with the following layer-wise propagation rule:
\begin{equation}
  \label{eqn:gcn}
  \mathbf{H}^{(l+1)} = \sigma\left( \tilde{\mathbf{G}}  \mathbf{H}^{(l)}\mathbf{W}^{(l)}\right).
\end{equation}
Here, $\tilde{\mathbf{G}} = \tilde{\mathbf{D}}^{-\frac{1}{2}}\tilde{\mathbf{A}}\tilde{\mathbf{D}}^{-\frac{1}{2}} = (\mathbf{D} + \mathbf{I}_{n})^{-\frac{1}{2}}(\mathbf{A} + \mathbf{I}_{n})(\mathbf{D} + \mathbf{I}_{n})^{-\frac{1}{2}}$ is the operator corresponding to the {\em Graph Convolution} in \cref{wide_and_deep}, where $\tilde{\mathbf{A}} = \mathbf{A} + \mathbf{I}_{n}$ is the adjacency matrix of the undirected graph $\mathcal{G}$ with added self-connections and $\tilde{\mathbf{D}}_{ii} = \sum_{j}\tilde{\mathbf{A}}_{ij}$.
$\mathbf{W}^{(l)}$ is a layer-specific trainable weight matrix corresponding to the {\em Linear Transformation} in \cref{wide_and_deep}.
$\sigma(\cdot)$ denotes the $ReLU(\cdot ) = max(0, \cdot)$. 
$\mathbf{H}^{(l)} \in \mathbb{R}^{n \times d}$ is the matrix of activations in the $l$-th layer and $\mathbf{H}^{(0)} = \mathbf{X}$ is the node feature matrix.

\paragraph{ResNet.}
The famous work ResNet from \citet{he16} solves the difficult problem of training deep neural networks in a very simple way called {\em residual connection}, which can be formalized as $\mathbf{y} = \mathcal{F}(\mathbf{x}) + \mathbf{x}$.
Inspired by \citet{he16}, \citet{kipf17} use {\em residual connection} between hidden layers to facilitate training of deeper GCN by enabling the model to carry over information from the previous layer’s input:
\begin{equation}
  \label{eqn:resgcn}
  \mathbf{H}^{(l+1)} = \sigma\left( \tilde{\mathbf{G}}\mathbf{H}^{(l)}\mathbf{W}^{(l)}\right) + \mathbf{H}^{(l)}.
\end{equation}
However, simply deepening the network does not bring additional benefits to GCN in node classification tasks.

\paragraph{APPNP.}
PPNP~\cite{gasteiger19} generates predictions for each node based on its own features and then propagates them via the fully personalized PageRank~\cite{page99} scheme to generate the final predictions.
PPNP's model is defined as $\mathbf{H} = \alpha(\mathbf{I}_{n}-(1-\alpha)\tilde{\mathbf{G}})^{-1} f_{\theta}(\mathbf{X})$, where $f_{\theta}(\cdot )$ denotes a 2-layer MLP.
\citet{gasteiger19} also proposes a fast approximation variant called APPNP with the following layer-wise propagation rule:
\begin{equation}
  \label{eqn:appnp}
  \mathbf{H}^{(l+1)} =(1-\alpha) \tilde{\mathbf{G}} \mathbf{H}^{(l)}+\alpha \mathbf{H}^{(0)},
\end{equation}
where $\mathbf{H}^{(0)} =f_{\theta}(\mathbf{X})$.
With this propagation rule, we can design very deep models even without using {\em residual connection} because there are no parameters in the graph convolution layer.
This provides a starting point for studying the strong generalization of {\em Graph Convolution} itself.

\paragraph{GCNII.}
\citet{chen20} propose the GCNII, an extension of the vanilla GCN model with two simple yet effective techniques: {\em Initial residual} and {\em Identity mapping}.
The idea of {\em Initial residual} is the same as the propagation rule of APPNP~\cite{gasteiger19} and {\em Identity mapping} is the concept proposed in \citet{he16}, which is a variant of {\em residual connection}.
Unlike \cref{eqn:resgcn}, the {\em identity mapping} in GCNII precedes the activation function, which is consistent with the design in ResNet~\cite{he16}.
Formally, GCNII's propagation rule is defined as:
\begin{equation}
  \label{eqn:gcnii}
  \hspace{-0.7mm} \mathbf{H}^{(l+1)} \hspace{-1.0mm}= \hspace{-1.0mm}
  \sigma  \hspace{-0.7mm}\left(  \hspace{-0.7mm}\left( \hspace{-0.7mm}  (1  \hspace{-0.7mm}-  \hspace{-0.7mm}\alpha_l)\tilde{\mathbf{G}}
  \mathbf{H}^{(l)}  \hspace{-0.7mm} +  \hspace{-0.7mm}
  \alpha_l \mathbf{H}^{(0)}  \hspace{-0.7mm}\right)  \hspace{-0.7mm}
  \left(  \hspace{-0.7mm}   (1  \hspace{-0.7mm} -  \hspace{-0.7mm}\beta_l) \mathbf{I}_n \hspace{-0.7mm} +
  \hspace{-0.7mm} \beta_l \mathbf{W}^{(l)}  \hspace{-0.7mm}\right)  \hspace{-0.7mm}\right),
\end{equation}
where $\beta_l = \lambda / l$ , $\alpha_l$ and $\lambda$ are two hyperparameters.
\section{GCNIII Model}
\label{sec:model}

We propose GCNIII, the first model to extend the Wide \text{\&} Deep learning to the field of graph-structured data, unifying the effective techniques from previous studies as hyperparameters.
We also propose the technical concept of embedding large language models (LLMs) into the framework for upgrading.
In all formulas below, $\{ \cdot \}$ represents non-essential module that need to be adjusted for different datasets and tasks.

\subsection{Wide \text{\&} Deep Learning}

\paragraph{The Wide Component.}
Generalized linear models with nonlinear feature transformations are widely used for large-scale regression and classification problems with sparse inputs.
When we encounter graph-structured data with sparse node features $\mathbf{X}\in \mathbb{R}^{n \times d}$, it is natural to wonder whether linear models can play a role in node classification.
We demonstrate the feasibility and validity of the linear models through solid experiments in \cref{sec:linear}.

Unlike linear regression model in \citet{cheng16}, the wide component here is a linear classification model of the form:
\begin{equation}
    \label{eqn:wide}
    \mathcal{W}(\mathbf{X}) = \{\psi \}(\mathbf{X}\mathbf{W}), 
\end{equation}
where $\mathbf{W} \in \mathbb{R}^{d \times c}$, $c$ is the number of categories of nodes and $\psi$ is the Batch Normalization~\cite{ioffe15}.
It should be emphasized that the generalization ability of linear models is extremely limited when the amount of data is small. 
Different from what people are familiar with, although Batch Normalization~\cite{ioffe15} can accelerate convergence and make the model have better classification ability in node classification tasks, it will reduce the generalization performance of the model.
We also provide a detailed analysis in \cref{sec:linear}.

\paragraph{The Deep Component.}
Compared with the feed-forward neural network in \citet{cheng16}, {\em Graph Convolution} can bring more amazing generalization ability improvement~\cite{yang23}.
The deep component is a simple yet flexible GCN model with two core components of {\em Graph Convolution} and {\em Linear Transformation}:
\begin{equation}
    \label{eqn:deep}
    \mathbf{H}^{(0)} = \{\sigma\}\left(\{\rho\} (\mathbf{X})\mathbf{W}_{\mathbf{e}}\right),
\end{equation}

\vspace{-5mm}

\begin{equation}
    \label{eqn:deep1}
    \mathbf{H}^{(l+1)} = \{\sigma\}\left(\{\tau \}(\tilde{\mathbf{G}}) \{\rho \}(\mathbf{H}^{(l)})\{\mathbf{W}^{(l)}\}\right),
\end{equation}

\vspace{-5mm}

\begin{equation}
    \label{eqn:deep2}
    \mathcal{D}(\mathbf{A}, \mathbf{X}) = \{\rho \} (\mathbf{H}^{(L)})\mathbf{W}_{\mathbf{p}}.
\end{equation}

$\mathbf{H}^{(L)}$ is the final layer of propagation. $\mathbf{W}_{\mathbf{e}}$ is the parameter matrix for dimensionality reduction of node features corresponding to {\em Feature Embedding} in \cref{wide_and_deep} and $\mathbf{W}_{\mathbf{p}}$ is {\em Prediction Layer}.
$\rho(\cdot)$ and $\tau(\cdot)$ are Dropout~\cite{geoffrey14} and DropEdge~\cite{rong20}.

\paragraph{Joint Training of Wide \text{\&} Deep Model.}
We intend for the two components to be relatively independent, which means that their memorization and generalization abilities are not intertwined, so that we can better understand the sources of model improvement or degradation.
Therefore, the output of the model is:
\begin{equation}
    \label{eqn:output}
    \mathcal{P} = Softmax\left( \gamma \mathcal{W}(\mathbf{X}) + (1 -\gamma) \mathcal{D}(\mathbf{A}, \mathbf{X}) \right),
\end{equation}
where $\mathcal{P}_{i}$ represents the predicted class distribution for the $i$-th node.
Then we use the cross-entropy loss function and the Adam optimizer~\cite{kingma15} for joint training.

Although the memorization of the linear models is beneficial for recommendation systems, it requires a large amount of training data as support. 
When using the wide component for node classification, we should adjust the hyperparameter $\gamma$ based on the proportion of training set in the datasets and the characteristics of classification tasks, which has strong skills.
Moreover, we find that $\gamma$ cannot be trained as a parameter because the desired generalization ability of the model might not align with the reduction of the loss function value.

\subsection{Techniques as Hyperparameters}

\paragraph{Intersect memory.}
We are concerned that when the training data is limited, the poor generalization of the wide component may negatively impact the overall performance of the model. 
To address this, we propose a technique called {\em Intersect memory}. 
The output of the wide component is the distribution of nodes' categories, and we apply a prior attention transformation to this distribution:
\begin{equation}
    \label{eqn:im_wide}
    \mathcal{W}(\mathbf{A}, \mathbf{X}) = \mathbf{A}_{\mathbf{IM}}(\{\psi \}(\mathbf{X}\mathbf{W})) = \tilde{\mathbf{G}}(\{\psi \}(\mathbf{X}\mathbf{W})). 
\end{equation}
The attention matrix between the nodes is the adjacency matrix, allowing this process to be directly performed using {\em Graph Convolution}.
The improved model is still a linear model, but whether to use this technique depends on the datasets.

\paragraph{Initial residual.}
The prototype of this technique first appeared in \citet{gasteiger19}, inspired by personalized PageRank~\cite{page99}, where the authors defined the propagation rule given by \cref{eqn:appnp}.
From the perspective of {\em residual connection}~\cite{he16}, \citet{chen20} name this technique {Initial residual} as an improvement to the common residual that carries the information from the previous layer.
In this paper, we emphasize that this technique is particularly effective in overcoming {\em over-smoothing} when designing deep GCNs. 
We all know {\em over-smoothing} can degrade model performance, but the difficult-to-train parameters are the root cause of the sudden performance drop as GCNs deepen. 
Empirical evidence is presented in \cref{sec:linear_transformation}, which supports the viewpoint proposed in \citet{zhang21}.

\paragraph{Identity mapping.}
To design the deep GCN model, we need to apply the {\em Initial residual} technique and remove the parameters $\mathbf{W}^{(l)}$ from ~\cref{eqn:deep1}. 
However, removing the parameters will inevitably result in the loss of some information. 
To address this issue, the {\em Identity mapping} proposed in ResNet~\cite{he16} can alleviate the challenges of parameter optimization in deep networks.
The technique is used in a manner consistent with \cref{eqn:gcnii} rather than \cref{eqn:resgcn}.
It is important to note that {\em Identity mapping} can indeed provide a performance boost in deep GCNs, but the boost is relatively small compared to the large number of parameters added. 
% Therefore, it is important to weigh the advantages and disadvantages of this technique before using it.
In GCNIII, {\em Identity mapping} is an integral component, and {\em Linear Transformation} is not applied if this technique is not utilized.

\subsection{Feature Engineering with LLMs}
Although the strong generalization ability of {\em Graph convolution} is the main reason for GCNs' superior performance in node classification tasks, we emphasize the role of node features in this paper, with detailed experiments presented in \cref{sec:node_features}.

\begin{figure}[ht]
    \vspace{-5mm}
    \begin{center}
    \centerline{\includegraphics[width=8cm]{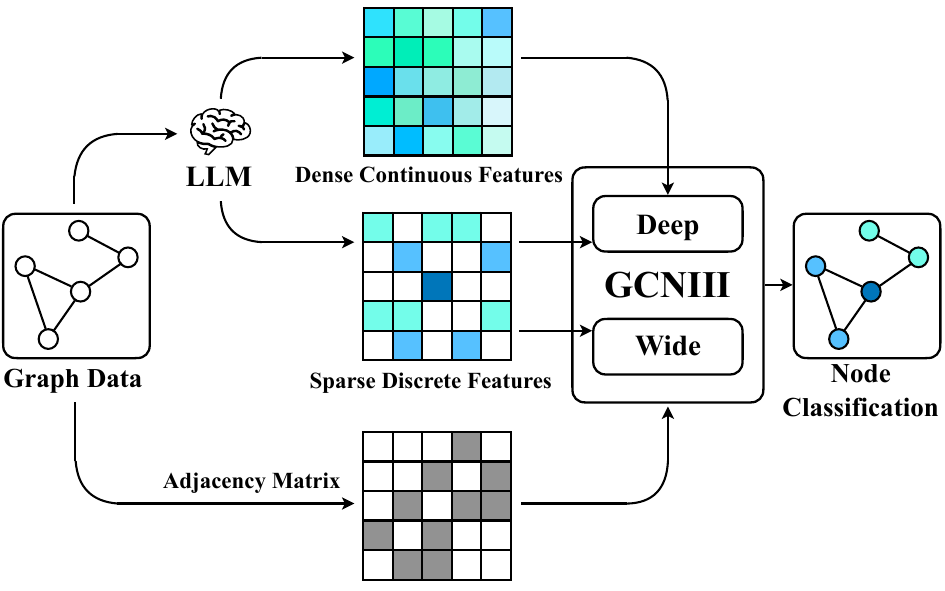}}
    \caption{LLM for GCNIII. 
    Sparse discrete features of graph nodes can be constructed using LLM, such as bag-of-words representation of document, which can be used in both the Wide and Deep Components. 
    A unified text-attribute description format can also be used to construct text-attribute graphs (TAGs) as input to the LLM, generating dense continuous features that enhance the learning of the Deep Component.}
    \label{llm_gcniii}
    \end{center}
    \vspace{-10mm}
\end{figure}

In the field of node classification tasks, many well-known datasets used in academic research have node characteristics carefully designed by the original authors.
When we want to apply the GCNIII model to graph data in other academic or industrial fields, recent research~\cite{he24} suggests that large language models (LLMs) may be efficient feature encoders, i.e. $\mathbf{X}=\mathbf{LLM}(\mathcal{G})$.
Cross-domain graph datas can even be encoded into the same embedding space using unified text-attribute graphs (TAGs)~\cite{liu24a}, further enhancing the potential of GCNIII as a pre-training foundation model for graphs.
The technical concept is illustrated in \cref{llm_gcniii}. % with the details provided in \cref{sec:llm}.

\section{Over-Generalization}
\label{sec:overgen}

\begin{figure}[ht]
    \begin{center}
    \centerline{\includegraphics[width=\columnwidth]{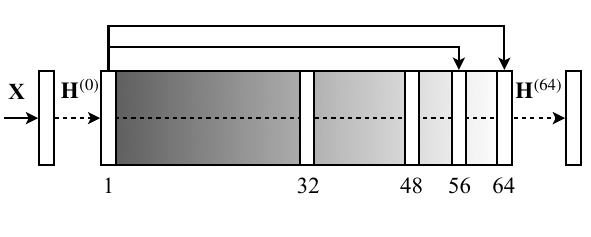}}
    \vspace{-5mm}
    \caption{Example network architecture for 64-layer GCNII. The color gradient from black to white represents the weight $\beta_l$ of the {\em Linear Transformation} from large to small.
    {\em Initial residual} inputs $\mathbf{H}^{(0)}$ directly to each layer, and the network between layers 56 and 64 contains an 8-layer sub-GCNII.
    This structure is similar to a reversed JKNet.}
    \label{gcnii}
    \end{center}
    \vspace{-7mm}
\end{figure}

To this day, GCNII remains one of the most outstanding deep GCN models.
\citet{chen20} prove that a $K$-layer GCNII can express a $K$ order polynomial filter $\left(\sum_{\ell=0}^{K}\theta_{\ell} \tilde{\mathbf{L}}^{\ell}\right) \mathbf{x}$ with arbitrary coefficients $\theta$, and this is considered a theoretical explanation for the superior performance of GCNII.
When training a 64-layer GCNII on the classic citation dataset Cora, we observe an uncommon phenomenon, as shown in \cref{train_val_error}.
We call this phenomenon {\em over-generalization}, which piques our interest in exploring the cause behind it and revisiting the source of GCNII's SOTA performance.

\paragraph{Dropout is the key.}
In fact, this can be easily inferred intuitively from \cref{train_val_error}, because dropout is the only component in the entire end-to-end GCNII model that has a different structure during training and inference.
Due to GCNII's complex structure, the authors~\cite{chen20} also do not find that dropout has such a significant impact on model performance.
Taking the Cora dataset as an example, we find through experimental studies that removing all dropout from GCNII results in a drop in accuracy from over 85\% to 82\%.
This means that although deep GCNII is theoretically capable of resolving the {\em over-smoothing} issue, without dropout, its actual performance is no different from a 2-layer GCN.
We also find the most critical of all dropout layers is the one before the {\em Feature Embedding} (shown in \cref{wide_and_deep}), and removing the dropout at this position will lead to a noticeable decrease in the model's accuracy.
Dropout in \cref{eqn:deep} is not commonly seen in the design of GCNs, but it is indeed one of the key aspects of the GCNII model.
\citet{geoffrey14} propose dropout as a regularization method by sampling from an exponential number of different “thinned” networks.
We argue that the dropout in \cref{eqn:deep} is more akin to a robust feature selection process, where a subset of features is randomly selected for feature embedding at each epoch. 
This process enhances the model's ability to efficiently leverage node feature information, thereby improving its generalization performance.

\paragraph{Ultra-deep is not necessary.}
Simply using dropout is not enough. 
In the right part of \cref{train_val_error}, the training error remains much higher than the validation error throughout the training process, while the validation error decreases very quickly.
For a 2-layer GCN, no matter how the dropout rate is set, {\em over-generalization} cannot occur. 
We propose that a certain model depth is a necessary condition for {\em over-generalization}, but how deep should GCNII be?
We find that an 8-layer GCNII removing {\em Identity mapping}, which is a variant of APPNP, is sufficient to achieve an average accuracy of 85\%.
This model has significantly fewer parameters compared to the original 64-layer GCNII, leading to a noticeable improvement in training speed. 
% If extreme performance is not the primary goal, this model seems to be a more cost-effective choice.
Our analysis suggests that unlike other deep neural networks, GCNII's power is primarily derived from the layers near the output.
As shown in \cref{gcnii}, the network from layers 56 to 64 contains the 8-layer GCNII described above, as the $\beta_{l}$ of these layers is close to 0.
To better understand the effect of model layers, we have the following Theorem.
\begin{theorem}
    \label{thm:deep}
    Let the $K$-layer GCNII model be $f_{K}(\mathbf{A}, \mathbf{X})$.
    $\forall \epsilon > 0$, $\exists K_{0} \in \mathbb{N}^{*}$ such that when $K > K_{0}$, we have
    ${\Vert f_{K+1}(\mathbf{A}, \mathbf{X}) - f_{K}(\mathbf{A}, \mathbf{X}) \Vert}_{2} < \epsilon$.
\end{theorem}
The proof of \cref{thm:deep} is in \cref{sec:proof}.
We also find that GCNII cannot contain a linear model, that is, feature information must pass through at least one two-layer MLP with ReLU activation from input to output, which is the motivation for our proposed GCNIII model.

\paragraph{Attention is all you need.}
We suggests that graph can be viewed as a form of static, discrete self-attention mechanism ~\cite{vaswani17}.
The matrix operation form of self-attention is:
\begin{equation}
  \label{eqn:attention}
  \text{softmax}\left( \frac{(\mathbf{X}\mathbf{W}^{Q})({\mathbf{X}\mathbf{W}^{K}})^{T}}{\sqrt{d_{k}}}  \right) \ast   (\mathbf{X}\mathbf{W}^{V}).
\end{equation}
{\em Graph Convolution} $\tilde{\mathbf{G}}$ corresponds to the attention matrix on the left-hand side of \cref{eqn:attention}.
Regardless of {Identity mapping}, {\em Initial residual} causes the “attention” of GCNII to asymptotically approach $\alpha(\mathbf{I}_{n}-(1-\alpha)\tilde{\mathbf{G}})^{-1}$ as the number of layers increases indefinitely.
Moreover, {\em Identity mapping} enables the “attention” to fine-tune through data.
A conventional attention matrix is typically dense and captures global attention information between elements.
We calculate the attention density values for both on Cora with $\alpha = 0.1$, i.e., the proportion of non-zero elements in the attention matrix.
The former is 0.0018, while the latter is 0.8423, which demonstrates that GCNII's “attention” captures more information, leading to stronger generalization.
Through a comparative analysis of misclassified nodes, we also find that 64-layer GCNII has stronger out-of-distribution generalization ability than 2-layer GCN, as detailed in \cref{sec:ood}.

%We have the following Theorem to illustrate the extrapolation power of GCNII.
%\begin{theorem}
  %\label{thm:extrapolation}
%\end{theorem}
%The proofs of \cref{thm:deep} and \cref{thm:extrapolation} can be found in \cref{sec:proof}.
\section{Other Related Work}
\label{sec:related}

The research of Graph Neural Networks (GNNs) for node classification is still a hot topic in machine learning.
\citet{song23} propose ordering message passing into node representations by aligning a central node's rooted-tree hierarchy with its ordered neurons in specific hops.
\citet{pei24} indicate that the root cause of {\em over-smoothing} and {\em over-squashing} is information loss due to heterophily mixing in aggregation.
\citet{zheng24b} disentangle the graph homophily into label, structural, and feature homophily.
Exploration of Graph Transformers in node classification tasks is still ongoing, \citet{wu22} propose a Transformer-style model with kernerlized Gumbel-Softmax operator that decreases the complexity to linearity, then \citet{xing24} improve Graph Transoformer with collaborative training to prevent the {\em over-globalizing} problem while keeping the ability to extract valuable information from distant nodes.
The impact of the data cannot be ignored, \citet{liu24b} study the effect of class imbalance on node classification from a topological paradigm and \citet{luo24a} combine GNNs and MLP to efficiently implement Sharpness-Aware Minimization (SAM), enhancing performance and efficiency in Few-Shot Node Classification (FSNC) tasks.
In addition to regular offline learning, \citet{zheng24a} conduct online evaluation of GNNs to gain insights into their effective generalization capability to real-world unlabeled graphs under test-time distribution shifts.
\citet{chen24} use LLMs as feature encoders for node classification.
\section{Experiments}
\label{sec:exp}

In this section, we evaluate the performance of GCNIII on a wide variety of open graph datasets.
Although LLMs are powerful and have been the focus of recent deep learning research, they are not the focus of this paper.
To maintain fairness, we do not use LLMs in the experiments.
The hyperparameter details of all models are presented in \cref{sec:hyperparameters}.

\subsection{Dataset and Configuration Details.}
\paragraph{Dataset.}
We use all datasets used for evaluating GCNII~\cite{chen20} to evaluate GCNIII.
For semi-supervised node classification, we utilize three well-known citation network datasets Cora, Citeseer, and Pubmed~\cite{sen08}, where nodes symbolize documents and edges denote citation relationships. 
Each node's feature is represented by a bag-of-words representation of the document.
For full-supervised node classification, we use web networks Chameleon~\cite{rozemberczki21}, Cornell, Texas, and Wisconsin~\cite{pei20} in addition to the above three datasets, where nodes represent web pages and edges signify hyperlinks connecting them.
Similarly, the features of the nodes are derived from the bag-of-words representation of the respective web pages.
For inductive learning, we use Protein-Protein Interaction (PPI) networks which contains 24 graphs, where nodes and edges represent proteins and whether there is an interaction between two proteins.
Positional gene sets, motif gene sets and immunological signatures are used as features.
The node features of these graphs are all sparse and discrete, which are suitable for the wide component of GCNIII.
Statistics of the datasets are shown in \cref{dataset}.

\begin{table}[t]
    \caption{Dataset statistics.}
    \label{dataset}
    \vspace{2mm}
    \begin{tabular}{lrrrr}
    \toprule
    Dataset  & Nodes & Edges & Features & Classes\\
    \midrule
    Cora       &  2,708 &   5,429 & 1,433 & 7 \\
    Citeseer   &  3,327 &   4,732 & 3,703 & 6 \\
    Pubmed     & 19,717 &  44,338 & 500 & 3 \\
    Chameleon  &  2,277 &  36,101 & 2,325 & 4 \\
    Cornell    &    183 &    295  & 1,703 & 5 \\
    Texas      &    183 &    309  & 1,703 & 5 \\
    Wisconsin  &    251 &    499  & 1,703 & 5 \\
    PPI        & 56,944 & 818,716 & 50 & 121 \\
    \bottomrule
    \end{tabular}
    \vspace{-5mm}
\end{table}

\paragraph{Configuration.}
The experiments are conducted on a Linux server equipped with an Intel(R) Xeon(R) Gold 6240 CPU @ 2.60GHz, 256GB RAM and 3 NVIDIA A100-SXM4-40GB GPUs.
Because of the small size of the datasets, we only use a single GPU to train the models.
All models are implemented in PyTorch~\cite{paszke19} version 2.2.1, DGL~\cite{wang20} version 2.3.0 with CUDA version 12.1 and Python 3.12.7.

\subsection{Semi-Supervised Node Classification}
\paragraph{Dataset Splitting.}
In the semi-supervised node classification task, we conduct a conventional fixed split of training/validation/testing~\cite{yang16} on the Cora, Citeseer, and Pubmed datasets, with 20 nodes per class for training, 500 nodes for validation and 1,000 nodes for testing.
In this experiment, the number of training nodes is small, which can better evaluate the generalization ability of the models.

\begin{figure}[ht]
    \begin{center}
    % \vspace{-2mm}
    \centerline{\includegraphics[width=\columnwidth]{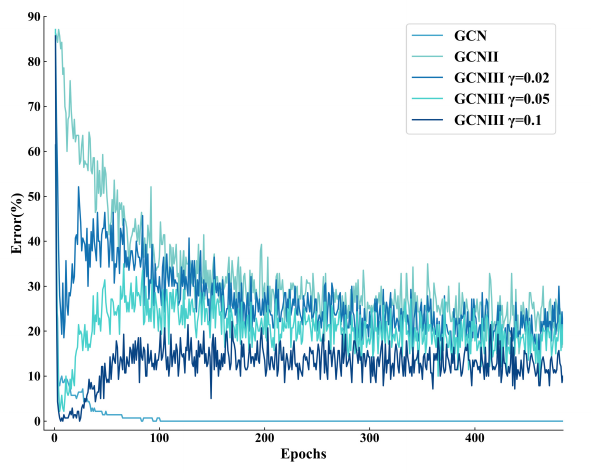}}
    \vspace{-5mm}
    \caption{Training error of the semi-supervised task on Cora with GCN, GCNII and GCNIII.}
    \label{train_errors}
    \end{center}
    \vspace{-10mm}
\end{figure}

\paragraph{Classic GNNs are Strong Baselines.}
\citet{luo24b} suggest that the performance of classic GNN models~\cite{kipf17,veli18,hamilton17} may be underestimated due to suboptimal hyperparameter configurations;
therefore, we use shallow SOTA models GCN~\cite{kipf17} and GAT~\cite{veli18}, as well as deep SOTA models APPNP~\cite{gasteiger19} and GCNII~\cite{chen20}, as baselines.
However, we do not directly reuse the metrics reported in \citet{luo24b} because we find that \citet{luo24b} use some unfair tricks in training. 
The open-source code of \citet{luo24b} shows that test accuracy is calculated and output for each training epoch, and the model accuracy result is the highest test accuracy of all epochs. 
\citet{luo24b} also randomly repartition the datasets to get “lucky” higher accuracy.
Therefore, we re-conduct the experiments in accordance with the optimal model hyperparameters reported in \citet{luo24b} under our experimental framework.
Details are presented in \cref{sec:records}.

\begin{table}[t]
    \caption{Accuracy (\%) results on Cora, Citeseer, and Pubmed in the semi-supervised node classification task. 
    The number in parentheses represents the number of layers in the model, and for GCNIII, it indicate the number of layers of the Deep Component model.}
    \label{semi_accuracy}
    \vspace{2mm}
    \setlength{\tabcolsep}{0.8mm}{
    \begin{tabular}{llll}
    \toprule
    Model & Cora & Citeseer & Pubmed \\
    \midrule
    GCN & 81.9 $\pm$ 0.6 (3) & 71.8 $\pm$ 0.1 (2) & 79.5 $\pm$ 0.3 (2)\\
    GAT & 80.8 $\pm$ 0.6 (3) & 69.3 $\pm$ 0.8 (3) & 78.4 $\pm$ 0.9 (2) \\
    APPNP & 83.3 $\pm$ 0.3 (8) & 71.8 $\pm$ 0.3 (8) & 80.1 $\pm$ 0.2 (8) \\
    GCNII & 85.2 $\pm$ 0.4 (64) & 72.8 $\pm$ 0.6 (32) & 79.8 $\pm$ 0.4 (16) \\
    \midrule
    GCNIII & \textbf{85.6 $\pm$ 0.4} (64) & \textbf{73.0 $\pm$ 0.5} (16) & \textbf{80.4 $\pm$ 0.4} (16) \\
    \bottomrule
    \end{tabular}}
    \vspace{-5mm}
\end{table}

\paragraph{Comparison with SOTA.}
The model implementation in \citet{chen20} is based on the PyG library~\cite{fey19}. 
To eliminate the influence of PyG~\cite{fey19} and DGL~\cite{wang20} on the model performance, we reproduce the results of APPNP~\cite{gasteiger19} and GCNII~\cite{chen20} using the hyperparameters in \citet{chen20}.
We train all the models using the same early stopping method in \citet{chen20} for fairness.
\cref{semi_accuracy} reports the mean classification accuracy with the standard deviation of each model after 10 runs.
Each run we use a different random seed to ensure that the model is evaluated as fairly as possible.
Our experimental results show that GCNIII has improved on the basis of GCNII~\cite{chen20}, achieving new state-of-the-art the performance on all three datasets.

\paragraph{Over-Generalization of GCNIII} 
Using the Cora dataset as an example, \cref{train_errors} illustrates the training error curves of GCN, GCNII and GCNIII.
We believe that the improved training error curve of GCNIII indicates a better balance between the model's fitting ability and generalization, which successfully demonstrates that GCNIII can more effectively balance the trade-off between the {\em over-fitting} of GCN and the {\em over-generalization} of GCNII.
As $\gamma$ increases, this balance improves, but it cannot be too large, or it will still lead to {\em over-fitting}.

\subsection{Full-Supervised Node Classification}
Following \citet{chen20}, we evaluate GCNIII in the full-supervised node classification task with 7 datasets: Cora, Citeseer, Pubmed, Chameleon, Cornell, Texas, and Wisconsin.
\citet{pei20} first randomly split nodes of each class into 60\%, 20\%, and 20\% for training, validation and testing, and measure the performance of all models by the average performance on the test sets over 10 random splits.
\citet{chen20} follow the criteria for splitting the dataset, so we also adopt the same standard.
Besides the previously mentioned models, we also include three variants of Geom-GCN~\cite{pei20} as the baseline. 
We reuse the metrics already reported in \citet{chen20} for GCN, GAT, Geom-GCN-I, Geom-GCN-P, Geom-GCN-S and APPNP.

\begin{table}[t]
    % \vspace{-2mm}
    \caption{Micro-averaged F1 scores on PPI.}
    \label{ppi_f1}
    \vspace{-2mm}
    \begin{center}
    % \begin{small}
    \setlength{\tabcolsep}{3.3mm}{
    \begin{tabular}{ll}
    \toprule
    Model & PPI\\
    \midrule
    GraphSAGE~\cite{hamilton17} & 61.2 \\
    GAT~\cite{veli18} & 97.3 \\
    VR-GCN~\cite{chen18} & 97.8 \\
    Cluster-GCN~\cite{chiang19} & 99.36 \\
    GCNII~\cite{chen20} & 99.48 $\pm$ 0.04 \\
    \midrule
    GCNIII & \textbf{99.50} $\pm$ \textbf{0.03}\\
    \bottomrule
    \end{tabular}}
    % \end{small}
    \end{center}
    \vspace{-5mm}
\end{table}

\begin{table*}[t]
    \vspace{-2mm}
    \caption{Mean classification accuracy of full-supervised node classification.}
    \label{full_accuracy}
    \vspace{2mm}
    \begin{center}
    % \begin{small}
    \begin{tabular}{llllllll}
    \toprule
    Model & Cora & Cite. & Pumb. & Cham. & Corn. & Texa. & Wisc. \\
    \midrule
    GCN & 85.77 & 73.68 & 88.13 & 28.18 & 52.70 & 52.16 & 45.88 \\
    GAT & 86.37 & 74.32 & 87.62 & 42.93 & 54.32 & 58.38 & 49.41 \\
    Geom-GCN-I & 85.19 & \textbf{77.99} & \textbf{90.05} & 60.31 & 56.76 & 57.58 & 58.24 \\
    Geom-GCN-P & 84.93 & 75.14 & 88.09 & 60.90 & 60.81 & 67.57 & 64.12 \\
    Geom-GCN-S & 85.27 & 74.71 & 84.75 & 59.96 & 55.68 & 59.73 & 56.67 \\
    APPNP & 87.87 & 76.53 & 89.40 & 54.3 & 73.51 & 65.41 & 69.02 \\
    GCNII & 88.35 (64) & 77.11 (64) & 89.58 (64) & 54.4 (8) & 59.46 (16) & 65.68 (32) & 65.69 (16) \\
    \midrule
    GCNIII & \textbf{88.47} (8) & 77.33 (8) & 89.88 (32) & \textbf{64.69} (2) & \textbf{74.59} (2) & \textbf{79.73} (2) & \textbf{83.33} (3)\\
    \bottomrule
    \end{tabular}
    % \end{small}
    \end{center}
    \vspace{-9mm}
\end{table*}

Table~\ref{full_accuracy} reports the mean classification accuracy of each model.
We retrain GCNII within our experimental framework using the hyperparameter settings in \citet{chen20}, however, the results we get on the last four datasets are much lower than those reported in \citet{chen20}.
We observe that GCNIII outperforms GCNII on all 7 datasets, especially the last four {\em heterophily} datasets, highlighting the superiority of the Wide \text{\&} Deep GCNIII model. 
This result suggests that the introduction of the Wide Component linear model enhances GCNIII's predictive power, surpassing the deep-only GCNII model.

\subsection{Inductive Learning}
Both semi-supervised and full-supervised node classification tasks require that all nodes in the graph are present during training.
\citet{hamilton17} first propose the inductive learning, which aims to leverage node feature information to efficiently generate node embeddings for previously unseen data.
Following \citet{veli18}, we use the Protein-Protein Interaction (PPI) dataset for the inductive learning task, with 20 graphs for training, 2 graphs for validation and the rest for testing.
We compare GCNIII with the following models:
GraphSAGE~\cite{hamilton17},
GAT~\cite{veli18}
VR-GCN~\cite{chen18},
Cluster-GCN~\cite{chiang19},
and GCNII~\cite{chen20}. 
We reuse the metrics reported in \citet{chen20}, except for GCNII.
We re-evaluated GCNII within our experimental framework to better understand the role of the Wide Component in GCNIII.
Table~\ref{ppi_f1} indicates that GCNIII outperforms GCNII on PPI, which demonstrates that the memorization ability of the linear model in the Wide Component can also play a role in inductive learning task.

\begin{table}[t]
    \caption{Ablation study on the Wide Component, where {\em wide} stands for the linear model in the Wide Component, and the number after “+” indicates the improved accuracy.}
    \label{ablation_study_wide}
    \vspace{2mm}
    \setlength{\tabcolsep}{2.6mm}{
    \begin{tabular}{llll}
    \toprule
    Model & Cora & Citeseer & Pubmed \\
    \midrule
    GCN & 81.5 & 71.0 & 79.2 \\
    GCN +{\em wide} & 81.8 +0.3 & 71.9 +0.9 & 80.1 +0.9 \\
    GAT & 83.0 & 70.4 & 77.9 \\
    GAT +{\em wide} & 83.2 +0.2 & 70.7 +0.3 & 78.1 +0.2 \\
    APPNP & 83.4 & 71.4 & 79.9 \\
    APPNP +{\em wide} & 83.7 +0.3 & 71.5 +0.1 & 80.5 +0.6 \\
    \bottomrule
    \end{tabular}}
    \vspace{-5mm}
\end{table}

\subsection{Ablation Study}
\paragraph{Effect of the Wide Component.}

The three experiments above have confirmed the effectiveness of the Wide Component to some extent, but its initial purpose is to alleviate the {\em over-generalization} of deep GCNs, so it may not provide benefits to shallow models prone to {\em over-fitting}.
Therefore, we add a basic linear classification model to 2-layer GCN, 2-layer GAT, and 8-layer APPNP as Wide Componet, and set $\gamma = 0.1$.
We still perform semi-supervised training on the classical datasets to compare the models.
The results in \cref{ablation_study_wide} show that the Wide Component can still play a role in the shallow models.

\begin{table}[t]
    \caption{Ablation study on three techniques, where {\em memo} stands for {\em Intersect memory}, {\em res} stands for {\em Initial residual}, {\em map} stands for {\em Identity mapping} and “-” indicates that the technique is removed.}
    \label{ablation_study_techniques}
    \vspace{2mm}
    \setlength{\tabcolsep}{2.1mm}{
    \begin{tabular}{llll}
    \toprule
    Model & Cora & Citeseer & Pubmed \\
    \midrule
    GCNIII & 85.1 & 72.8  & 79.5 \\
    GCNIII -{\em memo}& 84.7 -0.4 & 73.8 +1.0 & 79.6 +0.1 \\
    GCNIII -{\em res}& 63.1 -22.0 & 29.5 -43.3 & 51.2 -28.3 \\
    GCNIII -{\em map}& 85.7 +0.6 & 72.7 -0.1 & 79.4 -0.1 \\
    \bottomrule
    \end{tabular}}
    \vspace{-7mm}
\end{table}

\paragraph{Effect of three techniques.}
\cref{ablation_study_techniques} presents the results from an ablation study, which assesses the individual contributions of our three techniques: {\em Intersect memory}, {\em Initial residual}, and {\em Identity mapping}.
To ensure fairness, we apply the same hyperparameter settings uniformly across the three datasets: $\alpha_{l}=0.1$, $\lambda=0.5$, $\gamma=0.1$, 64 layers, 64 hidden units, dropout rate of 0.5 and learning rate of 0.01.
In this experiment, we fixed the random seed as 42, so the results have a certain randomness. 
However, we can still conclude that {\em Initial residual}, as shown in \cref{eqn:appnp}, is the most influential factor for the deep GCNs with dropout applied at each layer, while the other two have destabilizing effects and should be applied specifically according to the dataset. 
Combined with previous experiments, these three techniques are generally beneficial when used properly.
\section{Conclusion}
\label{sec:conclusion}

In this paper, we find that the training error is much higher than the validation error during the training process when studying the deep GCNII model, and we refer to this phenomenon as {\em over-generalization}.
We conduct an in-depth analysis of this phenomenon and propose GCNIII, the first model to extend the Wide \text{\&} Deep architecture to graph data.
We provide theoretical and empirical evidence that the Wide \text{\&} Deep GCNIII model more effectively balances the trade-off between {over-fitting} and {over-generalization} and achieves state-of-the-art results on various node classification tasks.
One meaningful direction for future work is to achieve more efficient node feature representation and graph structure construction by combining LLMs and GCNs models.

% \section*{Acknowledgements}

% \textbf{Do not} include acknowledgements in the initial version of
% the paper submitted for blind review.

% If a paper is accepted, the final camera-ready version can (and
% usually should) include acknowledgements.  Such acknowledgements
% should be placed at the end of the section, in an unnumbered section
% that does not count towards the paper page limit. Typically, this will 
% include thanks to reviewers who gave useful comments, to colleagues 
% who contributed to the ideas, and to funding agencies and corporate 
% sponsors that provided financial support.

\section*{Impact Statement}
GCNIII is an extension of deep graph convolutional networks and does not infringe upon Google's Wide \text{\&} Deep model. 
The application of Large Language Models (LLMs) may raise concerns about transparency and fairness in automated decision-making, potentially exacerbating existing biases. 
However, we believe that these broader implications align with the ongoing development of AI technologies.


\bibliography{gcniii}
\bibliographystyle{icml2025}

%%%%%%%%%%%%%%%%%%%%%%%%%%%%%%%%%%%%%%%%%%%%%%%%%%%%%%%%%%%%%%%%%%%%%%%%%%%%%%%
%%%%%%%%%%%%%%%%%%%%%%%%%%%%%%%%%%%%%%%%%%%%%%%%%%%%%%%%%%%%%%%%%%%%%%%%%%%%%%%
% APPENDIX
%%%%%%%%%%%%%%%%%%%%%%%%%%%%%%%%%%%%%%%%%%%%%%%%%%%%%%%%%%%%%%%%%%%%%%%%%%%%%%%
%%%%%%%%%%%%%%%%%%%%%%%%%%%%%%%%%%%%%%%%%%%%%%%%%%%%%%%%%%%%%%%%%%%%%%%%%%%%%%%

\appendix
\appendix
\onecolumn
\section{Proof of \cref{thm:deep}}
\label{sec:proof}

As we explained in \cref{eqn:gcnii}, GCNII's core propagation rule is defined as:
\begin{equation}
    \label{eqn:a_1} 
    \mathbf{H}^{(l+1)} = \sigma \left(\left( (1 - \alpha_l)\tilde{\mathbf{G}}
    \mathbf{H}^{(l)} + \alpha_l \mathbf{H}^{(0)}\right)
    \left((1  - \beta_l) \mathbf{I}_n + \beta_l \mathbf{W}^{(l)}\right)\right).
\end{equation}
In the actual implementation of GCNII, $\alpha_{l} = \alpha \in (0, 1)$ is a constant.
To reduce the complexity of the notation, we define $\mathbf{W}_{\mathbf{I}}^{(l)} =  (1  - \beta_l) \mathbf{I}_n + \beta_l \mathbf{W}^{(l)}$, {\em Feature Embedding} parameter as $\mathbf{W}_{\mathbf{e}}$ and {\em Prediction Layer} parameter as $\mathbf{W}_{\mathbf{p}}$.
As $l \rightarrow \infty$, $\mathbf{W}_{\mathbf{I}}^{(l)} \rightarrow \mathbf{I}_n$, since $\beta_l = \frac{\lambda}{l} \rightarrow 0$.
ReLU operation $\sigma$ in \cref{eqn:a_1} is difficult to handle in analysis. We adopt the same assumption as in \citet{chen20}, that is, the input node feature vectors are all non-negative.
Furthermore, we may assume that the parameters of each layer can map non-negative inputs to non-negative outputs.
Therefore, we remove ReLU operation in the subsequent analysis, and the simplified propagation rule is:
\begin{equation*}
    \mathbf{H}^{(l+1)} = \left( (1 - \alpha)\tilde{\mathbf{G}}
    \mathbf{H}^{(l)} + \alpha \mathbf{H}^{(0)}\right)\mathbf{W}_{\mathbf{I}}^{(0)}.
\end{equation*}
Initial $\mathbf{H}^{(0)} = \mathbf{X}\mathbf{W}_{e}$ is propagated layer by layer
\begin{equation*}
    \mathbf{H}^{(1)} = (1 - \alpha)\tilde{\mathbf{G}}\mathbf{H}^{(0)}\mathbf{W}_{\mathbf{I}}^{(0)} + \alpha \mathbf{H}^{(0)}\mathbf{W}_{\mathbf{I}}^{(1)},
\end{equation*}
\begin{equation*}
    \mathbf{H}^{(2)} = (1 - \alpha)^{2}\tilde{\mathbf{G}}^{2}\mathbf{H}^{(0)}\mathbf{W}_{\mathbf{I}}^{(0)}\mathbf{W}_{\mathbf{I}}^{(1)} + (1 - \alpha)\alpha\tilde{\mathbf{G}}\mathbf{H}^{(0)}\mathbf{W}_{\mathbf{I}}^{(0)}\mathbf{W}_{\mathbf{I}}^{(1)} + \alpha\mathbf{H}^{(0)}\mathbf{W}_{\mathbf{I}}^{(1)},
\end{equation*}
\begin{equation*}
    \dots\dots
\end{equation*}
Assuming the model has $K$ layers, we can express the final representation as:
\begin{equation}
    \label{eqn:a_2}
    \mathbf{H}^{(K)} = (1 - \alpha)^{K}\tilde{\mathbf{G}}^{K}\mathbf{H}^{(0)}\prod_{l=0}^{K-1}\mathbf{W}_{\mathbf{I}}^{(l)} + \alpha\sum_{i = 0}^{K-1}\left((1-\alpha)^{i}\tilde{\mathbf{G}}^{i}\mathbf{H}^{(0)}\prod_{k=K-i-1}^{K-1}\mathbf{W}_{\mathbf{I}}^{(k)}\right).
\end{equation}
In \cref{sec:overgen}, we analysis 64-layer GCNII with $\alpha = 0.1$, then $(1 - \alpha)^{64}\approx 0.001$.
This means that the 64-layer model represented by the first part on the right of \cref{eqn:a_2} hardly works, and parameters with so many layers are difficult to optimize using the back-propagation algorithm~\cite{hinton86}.
From the second part on the right of \cref{eqn:a_2}, we can find that the smaller $i$ is, the larger $(1-\alpha)^{i}$ becomes, the deeper parameter layer is, and the closer these parameters are to $\mathbf{I}_n$.

Surprisingly, the GCNII model explicitly combines all $k$-layer GCNs ($k=1,2,...,64$), but it is the shallow models that first come into play, and these models are actually located in the deeper layers, closer to the output, rather than near the input.
The above analysis focuses on the inference phase of the model; however, during the training phase of the model, the impact of dropout cannot be ignored, and we consider it a direction for future research.

We assume that $f_{K}(\mathbf{A}, \mathbf{X})$ and $f_{K+1}(\mathbf{A}, \mathbf{X})$ share the same parameter $\mathbf{W}_{\mathbf{e}}$ and $\mathbf{W}_{\mathbf{p}}$; otherwise, the randomness and complexity of the parameters would inevitably introduce errors.
In fact, the focus of our analysis here s the depth of the model, so this assumption is relatively reasonable.
${\Vert \mathbf{A}  \Vert}_{2} = (\lambda_{\mathbf{A}^{T}\mathbf{A}})^{\frac{1}{2}}$ is the $l_{2}$-induced norm or spectral norm, where $\lambda_{\mathbf{A}^{T}\mathbf{A}}$ denotes the largest eigenvalue of $\mathbf{A}^{T}\mathbf{A}$.
It's very easy to prove that ${\Vert \cdot  \Vert}_{2} $ is a consistent matrix norm, i.e., ${\Vert \mathbf{A}\mathbf{B} \Vert}_{2} \leqslant {\Vert \mathbf{A} \Vert}_{2} {\Vert \mathbf{B} \Vert}_{2}$.
We also assume that the width of the model is limited. To be more precise, $\mathbf{W}_{\mathbf{I}}^{(l)} \in \mathbb{R}^{n \times n}$ and $n$ is typically in the range of tens or hundreds in practical implementations.
In the following, we use $\mathbf{I}$ instead of $\mathbf{I}_{n}$.

The spectral norm is the maximum singular value of a matrix, and for a symmetric matrix, the spectral norm is equal to the absolute value of its largest eigenvalue.
$\tilde{\mathbf{G}} = \tilde{\mathbf{D}}^{-\frac{1}{2}}\tilde{\mathbf{A}}\tilde{\mathbf{D}}^{-\frac{1}{2}} = (\mathbf{D} + \mathbf{I})^{-\frac{1}{2}}(\mathbf{A} + \mathbf{I})(\mathbf{D} + \mathbf{I})^{-\frac{1}{2}}$ is a symmetric positive semidefinite matrix.
The maximum eigenvalue of $\mathbf{A} + \mathbf{I}$ does not exceed $\text{max}(\mathbf{D}_{ii}+1)$, while ${(\mathbf{D} + \mathbf{I})}^{-\frac{1}{2}}$ normalizes the eigenvalue to the range $[0, 1]$, then it is easy to deduce that ${\Vert \tilde{\mathbf{G}} \Vert}_{2} \leqslant 1$.

\begin{theorem}~\cite{wu19}\label{thm:Lapla_eig_shrink} 
Let $\mathbf{A}$ be the adjacency matrix of an undirected, weighted, simple graph $\mathcal{G}$ without isolated nodes and with corresponding degree matrix $\mathbf{D}$. Let $\tilde{\mathbf{A}} = \mathbf{A} + \gamma \mathbf{I}$, such that $\gamma > 0$, be the augmented adjacency matrix with corresponding degree matrix $\tilde{\mathbf{D}}$. 
Also, let $\lambda_1$ and $\lambda_n$ denote the smallest and largest eigenvalues of $\boldsymbol{\Delta}_{\text{sym}}=\mathbf{I} -  \mathbf{D}^{-\frac{1}{2}}\mathbf{A}\mathbf{D}^{-\frac{1}{2}}$; similarly, let $\tilde{\lambda}_1$ and $\tilde{\lambda}_n$ be the smallest and largest eigenvalues of $\tilde{\boldsymbol{\Delta}}_{\text{sym}}=\mathbf{I} -  \tilde{\mathbf{D}}^{-\frac{1}{2}}\tilde{\mathbf{A}}\tilde{\mathbf{D}}^{-\frac{1}{2}}$. 
We have that
\begin{equation}
    0 = \lambda_1 = \tilde{\lambda}_1 < \tilde{\lambda}_n < \lambda_n. \label{eq:corollary_aug_laplacian}
\end{equation}
\end{theorem}

Using the properties of the {\em Rayleigh quotient}, we can easily prove that the range of eigenvalues of $\tilde{\mathbf{L}} = \mathbf{I} - \tilde{\mathbf{G}}$ is $[0, 2]$, and combining with the \cref{thm:Lapla_eig_shrink} proposed by \citet{wu19}, we can further get ${\Vert \mathbf{I} - \tilde{\mathbf{G}} \Vert}_{2} < 2$.

In the Adam optimizer~\cite{kingma15}, the regularization term $\lambda_{w}{\Vert \Theta \Vert}^{2}$ for weight decay is added to the loss function $\mathcal{L} $ to encourage the model to use smaller weights, thereby reducing overfitting.
Therefore, we assume that all parameters have a upper bound, i.e., ${\Vert \mathbf{W} \Vert}_{2} < c_{1}$.
Due to the existence of weight decay, the model parameters cannot grow indefinitely, and sparsity characteristics will emerge. 
Since $\lim_{k \to \infty} \mathbf{W}_{\mathbf{I}}^{(k)} = \mathbf{I}$, we further impose a stronger assumption that the product of any number of {\em Identity mapping} parameters is bounded above, i.e., $\prod_{i \leqslant k \leqslant j}\mathbf{W}_{\mathbf{I}}^{(k)} < C$.
We use $\mathbf{W}_{\mathbf{I}}^{(l)}$ and $\tilde{\mathbf{W}}_{\mathbf{I}}^{(l)}$ to represent the {\em Identity mapping} parameters of the layers in $f_{K}(\mathbf{A}, \mathbf{X})$ and $f_{K+1}(\mathbf{A}, \mathbf{X})$, respectively. 
It is important to emphasize that these parameters are misaligned equality, i.e., $\mathbf{W} _{\mathbf{I}}^{(l)} = \tilde{\mathbf{W}} _{\mathbf{I}}^{(l+1)}$, as our analysis above shows that GCNII is primarily influenced by the layers closer to the output.
The input node feature matrix is usually very sparse, and even for dense matrix, each row is normalized so we assume that $X$ also has a small upper bound, i.e., ${\Vert \mathbf{X} \Vert}_{2} < c_{2}$. 
Furthermore, we have ${\Vert \mathbf{H}^{(0)} \Vert}_{2} \leqslant {\Vert \mathbf{X} \Vert}_{2} \cdot {\Vert \mathbf{W}_{\mathbf{e}} \Vert}_{2} < c_{1}c_{2}$.

\begin{proof}
    By \cref{eqn:a_2}, we obtain:
    \begin{equation*}
        \begin{aligned}
            \mathbf{H}^{(K+1)} &= (1 - \alpha)^{K+1}\tilde{\mathbf{G}}^{K+1}\mathbf{H}^{(0)}\prod_{l=0}^{K}\tilde{\mathbf{W}}_{\mathbf{I}}^{(l)} + \alpha\sum_{i = 0}^{K}\left((1-\alpha)^{i}\tilde{\mathbf{G}}^{i}\mathbf{H}^{(0)}\prod_{k=K-i}^{K}\tilde{\mathbf{W}}_{\mathbf{I}}^{(k)}\right) \\
            &= (1 - \alpha)^{K+1}\tilde{\mathbf{G}}^{K+1}\mathbf{H}^{(0)}\prod_{l=0}^{K}\tilde{\mathbf{W}}_{\mathbf{I}}^{(l)} + \alpha(1 - \alpha)^{K}\tilde{\mathbf{G}}^{K}\mathbf{H}^{(0)}\prod_{l=0}^{K}\tilde{\mathbf{W}}_{\mathbf{I}}^{(l)} \\
            &+ \alpha\sum_{i = 0}^{K-1}\left((1-\alpha)^{i}\tilde{\mathbf{G}}^{i}\mathbf{H}^{(0)}\prod_{k=K-i}^{K}\tilde{\mathbf{W}}_{\mathbf{I}}^{(k)}\right).
        \end{aligned}
    \end{equation*}   
    We first consider:  
    \begin{equation*}
        \begin{aligned}
            &\Vert \alpha\sum_{i = 0}^{K-1}\left((1-\alpha)^{i}\tilde{\mathbf{G}}^{i}\mathbf{H}^{(0)} \prod_{k=K-i}^{K}\tilde{\mathbf{W}}_{\mathbf{I}}^{(k)}\right) -  \alpha\sum_{i = 0}^{K-1}\left((1-\alpha)^{i}\tilde{\mathbf{G}}^{i}\mathbf{H}^{(0)}\prod_{k=K-i-1}^{K-1}\mathbf{W}_{\mathbf{I}}^{(k)}\right) \Vert_{2} \\
            &\quad\quad\quad = \alpha\Vert\sum_{i = 0}^{K-1}\left((1-\alpha)^{i}\tilde{\mathbf{G}}^{i}\mathbf{H}^{(0)}(\prod_{k=K-i}^{K}\tilde{\mathbf{W}}_{\mathbf{I}}^{(k)} - \prod_{k=K-i-1}^{K-1}\mathbf{W}_{\mathbf{I}}^{(k)})\right) \Vert_{2} \\
            &\quad\quad\quad \leqslant \alpha\sum_{i = 0}^{K-1}(1-\alpha)^{i}{{\Vert \tilde{\mathbf{G}} \Vert}_{2}}^{i} \cdot {\Vert \mathbf{H}^{(0)} \Vert}_{2} \cdot  \Vert \prod_{k=K-i}^{K}\tilde{\mathbf{W}}_{\mathbf{I}}^{(k)} - \prod_{k=K-i-1}^{K-1}\mathbf{W}_{\mathbf{I}}^{(k)} \Vert_{2} = 0.
        \end{aligned}
    \end{equation*}
    Then we consider:  
    \begin{equation*}
        \begin{aligned}
            &\Vert (1 - \alpha)^{K+1}\tilde{\mathbf{G}}^{K+1}\mathbf{H}^{(0)}\prod_{l=0}^{K}\tilde{\mathbf{W}}_{\mathbf{I}}^{(l)} + \alpha(1 - \alpha)^{K}\tilde{\mathbf{G}}^{K}\mathbf{H}^{(0)}\prod_{l=0}^{K}\tilde{\mathbf{W}}_{\mathbf{I}}^{(l)} - (1 - \alpha)^{K}\tilde{\mathbf{G}}^{K}\mathbf{H}^{(0)}\prod_{l=0}^{K-1}\mathbf{W}_{\mathbf{I}}^{(l)}  \Vert_{2} \\
            &=\Vert (1 - \alpha)^{K+1}\tilde{\mathbf{G}}^{K+1}\mathbf{H}^{(0)}\prod_{l=1}^{K}\tilde{\mathbf{W}}_{\mathbf{I}}^{(l)}(\tilde{\mathbf{W}}_{\mathbf{I}}^{(0)} - \mathbf{I} + \mathbf{I}) + \alpha(1 - \alpha)^{K}\tilde{\mathbf{G}}^{K}\mathbf{H}^{(0)}\prod_{l=1}^{K}\tilde{\mathbf{W}}_{\mathbf{I}}^{(l)}(\tilde{\mathbf{W}}_{\mathbf{I}}^{(0)} - \mathbf{I} + \mathbf{I}) \\
            &- (1 - \alpha)^{K}\tilde{\mathbf{G}}^{K}\mathbf{H}^{(0)}\prod_{l=0}^{K-1}\mathbf{W}_{\mathbf{I}}^{(l)}  \Vert_{2} < (1 - \alpha)^{K+1}c_{1}c_{2}C{\Vert \tilde{\mathbf{W}}_{\mathbf{I}}^{(0)} - \mathbf{I} \Vert}_{2} + \alpha(1 - \alpha)^{K}c_{1}c_{2}C{\Vert \tilde{\mathbf{W}}_{\mathbf{I}}^{(0)} - \mathbf{I} \Vert}_{2} \\
            &+ (1 - \alpha)^{K+1}c_{1}c_{2}C {\Vert \tilde{\mathbf{G}} - I \Vert}_{2} < (1 - \alpha)^{K}c_{1}c_{2}C(c_{1} + 3 - 2\alpha).
        \end{aligned}
    \end{equation*}
    
    $\forall \epsilon$, let $(1 - \alpha)^{K}c_{1}c_{2}C(c_{1} + 3 - 2\alpha) = \epsilon/c_{1}$, we obtain $K = \frac{\log(\epsilon/c_{1}^{2}c_{2}C(c_{1}+3-2\alpha))}{\log(1-\alpha)}$.

    Set $K_0 = \left\lfloor \frac{\log(\epsilon/c_{1}c_{2}C(c_{1}+3-2\alpha))}{\log(1-\alpha)} \right\rfloor $, then when $K > K_{0}$, we have:
    \begin{equation*}
        \begin{aligned}
            {\Vert f_{K+1}(\mathbf{A}, \mathbf{X}) - f_{K}(\mathbf{A}, \mathbf{X}) \Vert}_{2} \leqslant {\Vert \mathbf{H}^{(K+1)} - \mathbf{H}^{(K)} \Vert}_{2}\cdot {\Vert \mathbf{W}_{\mathbf{p}} \Vert}_{2}  < \epsilon.
        \end{aligned}
    \end{equation*}
\end{proof}

\section{Hyperparameters Details}
\label{sec:hyperparameters}
Table~\ref{semi_hyperparameters} summarizes the training configuration of all model for semi-supervised. 
$L_{2_a}$ denotes the weight decay for {\em Feature Embedding} and {\em Prediction Layer}.
$L_{2_b}$ denotes the weight decay for {\em Linear Transformation}.
As defined in \citet{chen20},the relation between $\beta_{l}$ in \cref{eqn:gcnii} and $\lambda$ is $\beta_{l} = \lambda /l$.
The 0/1 in the list of techniques indicates whether {\em Intersect memory}, {\em Initial residual} and {\em Identity mapping} are used.

\begin{table*}[h]
    \vspace{-5mm}
    \caption{The hyperparameters for \cref{semi_accuracy}.}
    \label{semi_hyperparameters}
    \vspace{2mm}
    \begin{center}
    \begin{small}
        \setlength{\tabcolsep}{0.7mm}{
        \begin{tabular}{l|l|l}
        \toprule
            Dataset               & Model & Hyperparameters \\
        \midrule
            \multirow{5}{*}{Cora}   & GCN & \begin{tabular}[c]{@{}l@{}}layers: 3, lr: 0.001, hidden: 512, dropout: 0.7, $L_2$: 0.0005 \end{tabular} \\
                                    & GAT & \begin{tabular}[c]{@{}l@{}}layers: 3, lr: 0.001, hidden: 512, dropout: 0.2, $L_2$: 0.0005 \end{tabular} \\ 
                                    & APPNP & \begin{tabular}[c]{@{}l@{}}layers:8, lr: 0.01,  hidden: 64, $\alpha$: 0.1, dropout: 0.5, $L_2$: 0.0005 \end{tabular} \\ 
                                    & GCNII & \begin{tabular}[c]{@{}l@{}}layers: 64, lr: 0.01,  hidden: 64, $\alpha_\ell$: 0.1, $\lambda$: 0.5, dropout: 0.6, $L_{2_{a}}$: 0.01, $L_{2_{b}}$: 0.0005 \end{tabular} \\
                                    & GCNIII & \begin{tabular}[c]{@{}l@{}}layers: 64, lr: 0.01,  hidden: 64, $\alpha_\ell$: 0.1, $\lambda$: 0.5, $\gamma$: 0.02, dropout: 0.6, $L_{2_{a}}$: 0.01, $L_{2_{b}}$: 0.0005, techniques: [1, 1, 1] \end{tabular} \\
            \midrule
            \multirow{5}{*}{Citeseer}   & GCN & \begin{tabular}[c]{@{}l@{}}layers: 2, lr: 0.001, hidden: 512, dropout: 0.5, $L_2$: 0.0005 \end{tabular} \\
                                        & GAT & \begin{tabular}[c]{@{}l@{}}layers: 3, lr: 0.001, hidden: 256, dropout: 0.5, $L_2$: 0.0005 \end{tabular} \\ 
                                        & APPNP & \begin{tabular}[c]{@{}l@{}}layers: 8, lr: 0.01, hidden: 64, $\alpha$: 0.1, dropout: 0.5, $L_2$: 0.0005 \end{tabular} \\ 
                                        & GCNII & \begin{tabular}[c]{@{}l@{}}layers: 32, lr: 0.01, hidden: 256,  $\alpha_\ell$: 0.2, $\lambda$: 0.6, dropout: 0.7, $L_{2_{a}}$: 0.01, $L_{2_{b}}$: 0.0005 \end{tabular} \\
                                        & GCNIII & \begin{tabular}[c]{@{}l@{}}layers: 16, lr: 0.01, hidden: 256, $\alpha_\ell$: 0.1, $\lambda$: 0.5, $\gamma$: 0.01, dropout: 0.5, $L_{2_{a}}$: 0.01, $L_{2_{b}}$: 0.0005, techniques: [1, 1, 1] \end{tabular} \\ 
            \midrule
            \multirow{5}{*}{Pubmed}     & GCN & \begin{tabular}[c]{@{}l@{}}layers: 2, lr: 0.005, hidden: 256, dropout: 0.7, $L_2$: 0.0005 \end{tabular} \\
                                        & GAT & \begin{tabular}[c]{@{}l@{}}layers: 2, lr: 0.01, hidden: 512, dropout: 0.5, $L_2$: 0.0005 \end{tabular} \\ 
                                        & APPNP & \begin{tabular}[c]{@{}l@{}}layers: 8, lr: 0.01, hidden: 64, $\alpha$: 0.1, dropout: 0.5, $L_2$: 0.0005 \end{tabular} \\ 
                                        & GCNII & \begin{tabular}[c]{@{}l@{}}layers: 16, lr: 0.01, hidden: 256, $\alpha_\ell$: 0.1, $\lambda$: 0.4, dropout: 0.5, $L_{2_{a}}$ = $L_{2_{b}}$: 0.0005 \end{tabular} \\
                                        & GCNIII & \begin{tabular}[c]{@{}l@{}}layers: 16, lr: 0.01, hidden: 256, $\alpha_\ell$: 0.1, $\lambda$: 0.4, $\gamma$: 0.02, dropout: 0.5, $L_{2_{a}}$ = $L_{2_{b}}$: 0.0005, techniques: [1, 1, 1] \end{tabular} \\ 
        \bottomrule
        \end{tabular}}
    \end{small}
    \end{center}
    \vspace{-3mm}
\end{table*}

Table~\ref{full_hyperparameters} summarizes the training configuration of GCNIII for full-supervised. 

\begin{table*}[h]
    \vspace{-5mm}
    \caption{The hyperparameters for \cref{full_accuracy}.}
    \label{full_hyperparameters}
    \vspace{2mm}
    \begin{center}
    \begin{small}
        \setlength{\tabcolsep}{1.5mm}{
        \begin{tabular}{l|l|l}
        \toprule
            Dataset      & Model         & Hyperparameters \\
        \midrule
            \multirow{2}{*}{Cora}   & GCNII & \begin{tabular}[c]{@{}l@{}}layers: 64, lr: 0.01,  hidden: 64, $\alpha_{l}$: 0.2, $\lambda$: 0.5, dropout: 0.5, $L_{2_{a}}$ = $L_{2_{b}}$: 0.0001 \end{tabular} \\
                                    & GCNIII & \begin{tabular}[c]{@{}l@{}}layers: 8, lr: 0.01,  hidden: 64, $\alpha_{l}$: 0.2, $\lambda$: 0, $\gamma$: 0.02, dropout: 0.5, $L_{2_{a}}$ = $L_{2_{b}}$: 0.0001, techniques: [1, 1, 0] \end{tabular} \\
            \midrule
            \multirow{2}{*}{Citeseer}   & GCNII & \begin{tabular}[c]{@{}l@{}}layers: 64, lr: 0.01,  hidden: 64, $\alpha_{l}$: 0.5, $\lambda$: 0.5, dropout: 0.5, $L_{2_{a}}$ = $L_{2_{b}}$: 5e-6 \end{tabular} \\
                                        & GCNIII & \begin{tabular}[c]{@{}l@{}}layers: 8, lr: 0.01,  hidden: 128, $\alpha_{l}$: 0.5, $\lambda$: 1, $\gamma$: 0.02, dropout: 0.5, $L_{2_{a}}$ = $L_{2_{b}}$: 5e-6, techniques: [1, 1, 0] \end{tabular} \\
            \midrule
            \multirow{2}{*}{Pubmed} & GCNII & \begin{tabular}[c]{@{}l@{}}layers: 64, lr: 0.01,  hidden: 64, $\alpha_{l}$: 0.1, $\lambda$: 0.5, dropout: 0.5, $L_{2_{a}}$ = $L_{2_{b}}$: 5e-6 \end{tabular} \\
                                    & GCNIII & \begin{tabular}[c]{@{}l@{}}layers: 32, lr: 0.01,  hidden: 64, $\alpha_{l}$: 0.1, $\lambda$: 0.5, $\gamma$: 0.02, dropout: 0.6, $L_{2_{a}}$ = $L_{2_{b}}$: 5e-6, techniques: [1, 1, 1] \end{tabular} \\
            \midrule
            \multirow{2}{*}{Chameleon}  & GCNII & \begin{tabular}[c]{@{}l@{}}layers: 8, lr: 0.01,  hidden: 64, $\alpha_{l}$: 0.2, $\lambda$: 1.5, dropout: 0.5, $L_{2_{a}}$ = $L_{2_{b}}$: 0.0005 \end{tabular} \\
                                        & GCNIII & \begin{tabular}[c]{@{}l@{}}layers: 2, lr: 0.01,  hidden: 64, $\alpha_{l}$: 0, $\lambda$: 0, $\gamma$: 0.05, dropout: 0, $L_{2_{a}}$ = $L_{2_{b}}$: 0.0005, techniques: [1, 0, 0] \end{tabular} \\
            \midrule
            \multirow{2}{*}{Cornell}    & GCNII & \begin{tabular}[c]{@{}l@{}}layers: 16, lr: 0.01,  hidden: 64, $\alpha_{l}$: 0.5, $\lambda$: 1, dropout: 0.5, $L_{2_{a}}$ = $L_{2_{b}}$: 0.001 \end{tabular} \\
                                        & GCNIII & \begin{tabular}[c]{@{}l@{}}layers: 2, lr: 0.01,  hidden: 64, $\alpha_{l}$: 0.8, $\lambda$: 1, $\gamma$: 0.02, dropout: 0.5, $L_{2_{a}}$ = $L_{2_{b}}$: 0.001, techniques: [1, 1, 1] \end{tabular} \\
            \midrule
            \multirow{2}{*}{Texas}  & GCNII & \begin{tabular}[c]{@{}l@{}}layers: 32, lr: 0.01,  hidden: 64, $\alpha_{l}$: 0.5, $\lambda$: 1.5, dropout: 0.5, $L_{2_{a}}$ = $L_{2_{b}}$: 0.0001 \end{tabular} \\
                                    & GCNIII & \begin{tabular}[c]{@{}l@{}}layers: 2, lr: 0.01,  hidden: 64, $\alpha_{l}$: 0.5, $\lambda$: 1.5, $\gamma$: 0.05, dropout: 0.5, $L_{2_{a}}$ = $L_{2_{b}}$: 0.0001, techniques: [1, 1, 1] \end{tabular} \\
            \midrule
            \multirow{2}{*}{Wisconsin}  & GCNII & \begin{tabular}[c]{@{}l@{}}layers: 16, lr: 0.01,  hidden: 64, $\alpha_{l}$: 0.5, $\lambda$: 1, dropout: 0.5, $L_{2_{a}}$ = $L_{2_{b}}$: 0.0005 \end{tabular} \\
                                        & GCNIII & \begin{tabular}[c]{@{}l@{}}layers: 3, lr: 0.01,  hidden: 64, $\alpha_{l}$: 0.6, $\lambda$: 1, $\gamma$: 0.1, dropout: 0.8, $L_{2_{a}}$ = $L_{2_{b}}$: 0.0005, techniques: [1, 1, 1] \end{tabular} \\
            \bottomrule
            \end{tabular}}
    \end{small}
    \end{center}
    \vspace{-3mm}
\end{table*}

Table~\ref{ppi_hyperparameters} summarizes the training configuration of GCNIII for inductive learning.
Following \citet{veli18}, we add a skip connection from the $l$-th layer to the $(l + 1)$-th layer of GCNIII to speed up the convergence of the training process.

\begin{table*}[h]
    \vspace{-5mm}
    \caption{The hyperparameters for \cref{ppi_f1}.}
    \label{ppi_hyperparameters}
    \vspace{2mm}
    \begin{center}
    \begin{small}
        \setlength{\tabcolsep}{5.3mm}{
        \begin{tabular}{l|l}
        \toprule
            Model               & Hyperparameters \\
            \midrule
            GCNII & \begin{tabular}[c]{@{}l@{}}layers: 9, lr: 0.01,  hidden: 2048, $\alpha_\ell$: 0.5, $\lambda$: 1.0, dropout: 0.2, $L_{2_{a}}$: 0.0, $L_{2_{b}}$: 0.0  \end{tabular} \\
        
            GCNIII & \begin{tabular}[c]{@{}l@{}}layers: 9, lr: 0.01,  hidden: 2048, $\alpha_\ell$: 0.5, $\lambda$: 1.0, $\gamma$: 0.02, dropout: 0.2, $L_{2_{a}}$: 0.0, $L_{2_{b}}$: 0.0, techniques: [1, 1, 1] \end{tabular} \\
        \bottomrule
        \end{tabular}}
    \end{small}
    \end{center}
    \vspace{-3mm}
\end{table*}

It should be emphasized that we try to avoid using Dropedge in experiments, because dropedge will change the graph structure information during training, and this paper focuses on the role of node features.

\section{Linear Models for Node Classification.}
\label{sec:linear}
\citet{yang23} has confirmed the feasibility of MLPs~\cite{hinton86} for node classification tasks, so we wanted to explore the feasibility of linear models.
Since the goal of the task is to classify nodes, we explore the capability of a linear classification model in \cref{eqn:wide} with the most basic semi-supervised node classification task on Cora, Citeseer, and Pubmed Datasets.
We compare the linear classification model with the 2-layer MLP and 2-layer GCN, and explore adding Batch Normalization to these models.
We also evaluate the effect of {\em Intersect memory} technique on the linear model.
The effects of Dropout are unpredictable, and in order to facilitate a more intuitive comparison between linear model, MLP and GCN, Dropout is temporarily excluded from all models.
We conduct a fixed split of training/validation/testing~\cite{yang16} on the Cora, Citeseer, and Pubmed datasets, with 20 nodes per class for training, 500 nodes for validation and 1,000 nodes for testing.
For MLP and GCN, we fix the number of hidden units to 64 on all datasets and use ReLU as the activation function.
We train models using the Adam optimizer with a learning rate of 0.01 and $L_2$ regularization of 0.0005 for 200 epochs.
We report the final training loss(float), training accuracy(\%), and test accuracy(\%) in \cref{linear_semi_accuracy}, where BN represents Batch Normalization and IMLinear represents the linear classification model with {\em Intersect memory}.

\begin{table*}[h]
    \vspace{-3mm}
    \begin{center}
    \caption{Evaluation of the linear classification model in the semi-supervised node classification task.}
    \label{linear_semi_accuracy}
    \vspace{2mm}
    \begin{tabular}{l|ccc|ccc|ccc}
        \toprule
        Dataset & \multicolumn{3}{c|}{Cora} & \multicolumn{3}{c|}{Citeseer} & \multicolumn{3}{c}{Pubmed} \\
        \midrule
        Model & final loss & train acc & test acc & final loss & train acc & test acc & final loss & train acc & test acc\\
        \midrule
        Linear & 1.2971 & 100.0 & 54.0 & 1.3550 & 100.0 & 54.4 & 0.5797 & 100.0 & 71.6 \\
        Linear(+BN) & 0.0016 & 100.0 & 37.6 & 0.0008 & 100.0 & 39.7 & 0.0016 & 100.0 & 52.7 \\
        IMLinear & 1.5371 & 95.7 & 72.5 & 1.5514 & 95.8 & 65.6 & 0.7704 & 98.3 & 74.2 \\
        IMLinear(+BN) & 0.0074 & 100.0 & 63.4 & 0.0027 & 100.0 & 49.7 & 0.0086 & 100.0 & 60.1 \\
        MLP & 0.0772 & 100.0 & 59.7 & 0.0923 & 100.0 & 60.3 & 0.0329 & 100.0 & 72.6 \\
        MLP(+BN) & 0.0010 & 100.0 & 46.5 & 0.0009 & 100.0 & 47.6 & 0.0006 & 100.0 & 64.6 \\
        GCN & 0.1365 & 100.0 & 80.6 & 0.1831 & 100.0 & 71.5 & 0.0669 & 100.0 & 79.8 \\
        GCN(+BN) & 0.0016 & 100.0 & 76.4 & 0.0015 & 100.0 & 64.1 & 0.0009 & 100.0 & 75.2 \\
    \bottomrule
    \end{tabular}
    \end{center}
    \vspace{-3mm}
\end{table*}

We observe that the linear classification model can achieve 100\% accuracy on the training set, but the generalization ability is significantly weak, even a little weaker than 2-layer MLP.
{\em Intersect memory} can bring a large generalization ability to the linear classification model, while Batch Normalization can reduce the generalization ability of the models.
However, the number of nodes used for training in the semi-supervised task is too small to indicate the capability of the linear models. 
In general, when all nodes in the dataset are used to supervise training, the training accuracy is the upper limit of the model's ability. 
So we repeat the experiment above, but use all the nodes for training.

\begin{table*}[h]
    \vspace{-3mm}
    \begin{center}
    \caption{Evaluation of the linear classification model using all nodes for training.}
    \label{linear_all_accuracy}
    \vspace{2mm}
    \setlength{\tabcolsep}{5.3mm}{
    \begin{tabular}{l|cc|cc|cc}
        \toprule
        Dateset & \multicolumn{2}{c|}{Cora} & \multicolumn{2}{c|}{Citeseer} & \multicolumn{2}{c}{Pubmed} \\
        \midrule
        Model & final loss & train acc & final loss & train acc & final loss & train acc \\
        \midrule
        Linear & 1.5850 & 61.23 & 1.6185 & 71.81 & 0.8238 & 80.69 \\
        Linear(+BN) & 0.0428 & 100.00 & 0.0295 & 99.94 & 0.2859 & 89.80 \\
        IMLinear & 1.6286 & 61.41 & 1.6430 & 73.28 & 0.9049 & 79.49 \\
        IMLinear(+BN) & 0.0721 & 99.63 & 0.1063 & 97.14 & 0.3042 & 89.96 \\
        MLP & 0.2764 & 97.16 & 0.3953 & 93.12 & 0.3195 & 88.95 \\
        MLP(+BN) & 0.0031 & 100.00 & 0.0039 & 99.94 & 0.0088 & 99.99 \\
        GCN & 0.3554 & 92.10 & 0.5628 & 83.62 & 0.3619 & 87.59 \\
        GCN(+BN) & 0.0182 & 99.63 & 0.0672 & 96.93 & 0.0630 & 98.11 \\
    \bottomrule
    \end{tabular}}
    \end{center}
    \vspace{-3mm}
\end{table*}

From \cref{linear_all_accuracy}, we observe that the linear models have a poor classification ability when faced with more data, much lower than MLP.
But we accidentally find that Batch Normalization can improve this situation, and also improve the classification ability of MLP and GCN. 
Combining the results of the two groups of experiments, we basically verify that the linear models have the ability to handle the task of node classification. 
Batch Normalization can improve the classification ability of the models, but it will degrade the generalization ability, so it needs to be used selectively according to the actual situation.

\section{Linear Transformation of Deep GCNs.}
\label{sec:linear_transformation}
As is well known, {\em over-smoothing} is a major cause of the performance degradation in deep GCNs.
However, DNNs without graph convolution alse face challenges in parameter training as the number of layers increases. 
Each layer of the classical GCN model contains a parameterized {\em Linear Transformation}, and we aim to explore its impact on deep GCNs.
We still conduct the experiment on the semi-supervised node classification task, with the same setup in \cref{sec:linear}.
We compare GCN with its variant, GCN-v, which removes the linear transformation in each layer and retains only the graph convolution.
It is important to note that Dropout can improve the generalization ability of GCN to some extent; however, to more clearly observe the effect of {\em Linear Transformation} in the comparison experiment, we avoid using this technique.

\begin{table*}[h]
    \vspace{-3mm}
    \begin{center}
    \caption{Evaluation of {\em Linear Transformation} in deep GCN.}
    \label{lineartrans_deepgcn}
    \vspace{2mm}
    \setlength{\tabcolsep}{5mm}{
    \begin{tabular}{l|l|cccccccc}
        \toprule
        Dateset & Layer & 2 & 3 & 4 & 5 & 8 & 16 & 32 & 64 \\
        \midrule
        \multirow{2}{*}{Cora} & GCN & 80.5 & 80.3 & 75.4 & 71.8 & 57.4 & 27.0 & 27.2 & 24.2 \\
                              & GCN-v & 80.8 & 81.1 & 80.5 & 80.8 & 80.7 & 80.2 & 78.4 & 72.8\\
        \toprule
        \multirow{2}{*}{Citeseer} & GCN & 71.6 & 66.2 & 53.5 & 51.6 & 24.9 & 22.2 & 22.6 & 22.2 \\
                                  & GCN-v & 70.8 & 69.3 & 69.5 & 69.2 & 70.0 & 70.5 & 71.1 & 68.9 \\
        \toprule
        \multirow{2}{*}{Pubmed} & GCN & 80.0 & 78.4 & 75.2 & 74.4 & 63.0 & 44.1 & 40.1 & 42.8 \\
                                & GCN-v & 78.8 & 78.5 & 79.4 & 79.5 & 79.3 & 78.3 & 75.4 & 71.0 \\
    \bottomrule
    \end{tabular}}
    \end{center}
    \vspace{-3mm}
\end{table*}

\cref{lineartrans_deepgcn} reports the classification accuracy of GCN and GCN-v.
It is evident that the accuracy of GCN decreases sharply as the number of layers increases, while the accuracy of GCN-v without {\em Linear Transformation} remains stable until the number of layers reaches 32. 
Thus, we conclude that parameterized {\em Linear Transformation}, which is difficult to optimize, is the primary cause of the poor performance of deep GCNs, whereas {\em over-smoothing} plays a less significant role.

\section{Analysis of Node Features.}
\label{sec:node_features}
The input data for node classification typically consists of two components: the initial node features and the graph structure.
Most previous studies have focused on the graph structure, while we aim to explore the impact of node features on the performance of GCNs in node classification tasks.
We still follow the experimental settings in \cref{sec:linear} and use a 2-layer GCN for semi-supervised learning on the Cora, CiteSeer, and PubMed datasets. 
We will conduct comparative experiments using the following node features: 
1) Randomly generated features, which can be divided into binary discrete features and dense continuous features.
2) One-hot label encoding of the nodes, that is, the feature of the $i$-th node in graph $\mathcal{G}$ is a one-hot vector with the $i$-th position set to 1 and all other positions set to 0. 
The feature matrix of the entire graph $\mathcal{G}$ is the identity matrix $\mathbf{I} \in \mathbb{R}^{n \times n}$.
3) Learnable parameters, which are equivalent to the features obtained by applying a linear transformation of the same dimension to the one-hot vector features.
4) Bag-of-words representation of the nodes, which is a common initial feature for these datasets. These node features, proposed by \citet{sen08}, is very sparse and discrete. 
We also consider adding a dropout layer with 0.5 rate after this feature to verify the impact of randomly dropping some features during training on the model's performance.

\begin{table*}[h]
    \vspace{-3mm}
    \begin{center}
    \caption{Evaluation of the node features.}
    \label{node_features}
    \vspace{2mm}
    \setlength{\tabcolsep}{4mm}{
    \begin{tabular}{l|cc|cc|cc}
        \toprule
        Dataset & \multicolumn{2}{c|}{Cora} & \multicolumn{2}{c|}{Citeseer} & \multicolumn{2}{c}{Pubmed} \\
        \midrule
        Feature & dimension & accuracy & dimension & accuracy & dimension & accuracy \\
        \midrule
        Random(0-1) & 1000 & 54.2 & 1000 & 32.6 & 1000 & 36.3 \\
        Random(dense) & 1000 & 29.3 & 1000 & 28.3 & 1000 & 34.0 \\
        One-hot & 2708 & 63.1 & 3327 & 33.3 & 19717 & 38.2 \\
        Learnable parameters & 1000 & 57.4 & 1000 & 29.5 & 1000 & 33.3 \\
        Bag-of-words & 1433 & 80.7 & 3703 & 71.5 & 500 & 79.8 \\
        Bag-of-words(dropout) & 1433 & 82.1 & 3703 & 71.0 & 500 & 78.1 \\
    \bottomrule
    \end{tabular}
    }
    \end{center}
    \vspace{-3mm}
\end{table*}

The results in \cref{node_features} confirm that the quality of node features plays an extremely important role in the performance of GCN for node classification, which inspires us to explore constructing node features using powerful large language models(LLMs).

\section{Experimental Records.}
\label{sec:records}
The experimental results for the semi-supervised node classification tasks are detailed in \cref{semi_results}.
We use the hyperparameter settings reported in \citet{luo24b}, but do not achieve the same accuracy as in \citet{luo24b}, and the performance of GAT even degrades with these settings.

\begin{table*}[h]
    \vspace{-3mm}
    \caption{The results for \cref{semi_accuracy}.}
    \label{semi_results}
    \vspace{2mm}
    \begin{center}
        \setlength{\tabcolsep}{5.6mm}{
        \begin{tabular}{l|l|l|l|l}
        \toprule
            Dataset               & Mdoel & Results & Mean & Std \\
        \midrule
            \multirow{5}{*}{Cora}   & GCN & [81.9, 81.5, 81.5, 83.2, 82.9, 81.8, 81.0, 81.7, 81.8, 82.0] & 81.9 & 0.6 \\
                                    & GAT & [81.4, 80.6, 80.5, 81.6, 80.0, 80.6, 80.6, 81.4, 79.8, 81.0] & 80.8 & 0.6 \\ 
                                    & APPNP & [83.5, 83.3, 83.3, 83.3, 83.6, 83.6, 83.2, 83.3, 82.9, 82.8] & 83.3 & 0.3 \\ 
                                    & GCNII & [84.9, 85.9, 85.2, 84.9, 85.1, 84.8, 84.7, 85.4, 85.7, 85.1] & 85.2 & 0.4 \\
                                    & GCNIII & [85.6, 85.8, 84.9, 85.3, \textbf{86.1}, 84.9, 85.7, 85.6, 85.9, 85.7] & 85.6 & 0.4 \\
            \midrule
            \multirow{5}{*}{Citeseer}   & GCN & [71.7, 71.7, 71.7, 71.9, 72.0, 71.7, 71.6, 71.9, 71.9, 72.0] & 71.8 & 0.1 \\
                                        & GAT & [69.4, 70.3, 68.3, 69.2, 68.5, 69.8, 70.1, 68.5, 70.5, 68.5] & 69.3 & 0.8 \\ 
                                        & APPNP & [71.8, 72.0, 71.4, 71.6, 71.2, 72.0, 71.8, 72.5, 71.8, 71.5] & 71.8 & 0.3 \\ 
                                        & GCNII & [72.9, 73.3, 73.4, 71.7, 72.2, 72.8, 71.9, 72.7, 73.3, 73.4] & 72.8 & 0.6 \\
                                        & GCNIII & [73.2, 72.3, 72.9, 73.1, 73.0, 72.7, 72.3, \textbf{74.0}, 73.1, 73.5] & 73.0 & 0.5 \\ 
            \midrule
            \multirow{5}{*}{Pubmed}     & GCN & [80.0, 79.6, 79.3, 79.5, 79.3, 79.7, 79.0, 79.5, 79.3, 79.4] & 79.5 & 0.3 \\
                                        & GAT & [79.3, 78.1, 77.1, 77.7, 78.6, 80.2, 78.5, 79.2, 78.3, 77.3] & 78.4 & 0.9 \\ 
                                        & APPNP & [80.0, 79.9, 80.2, 80.4, 80.0, 80.3, 80.4, 80.1, 80.2, 79.9] & 80.1 & 0.2 \\ 
                                        & GCNII & [79.9, 79.3, 79.3, 80.1, 79.8, 80.1, 80.6, 80.0, 79.5, 79.8] & 79.8 & 0.4 \\
                                        & GCNIII & [80.0, 80.2, 80.0, \textbf{81.4}, 80.4, 80.6, 80.3, 80.6, 79.8, 80.5] & 80.4 & 0.4 \\ 
        \bottomrule
        \end{tabular}}
    \end{center}
    \vspace{-3mm}
\end{table*}

\section{Out-of-Distribution Generalization of GCNII.}
\label{sec:ood}

\begin{figure}[h]
    \vspace{3mm}
    \begin{center}
    \centerline{\includegraphics[width=10cm]{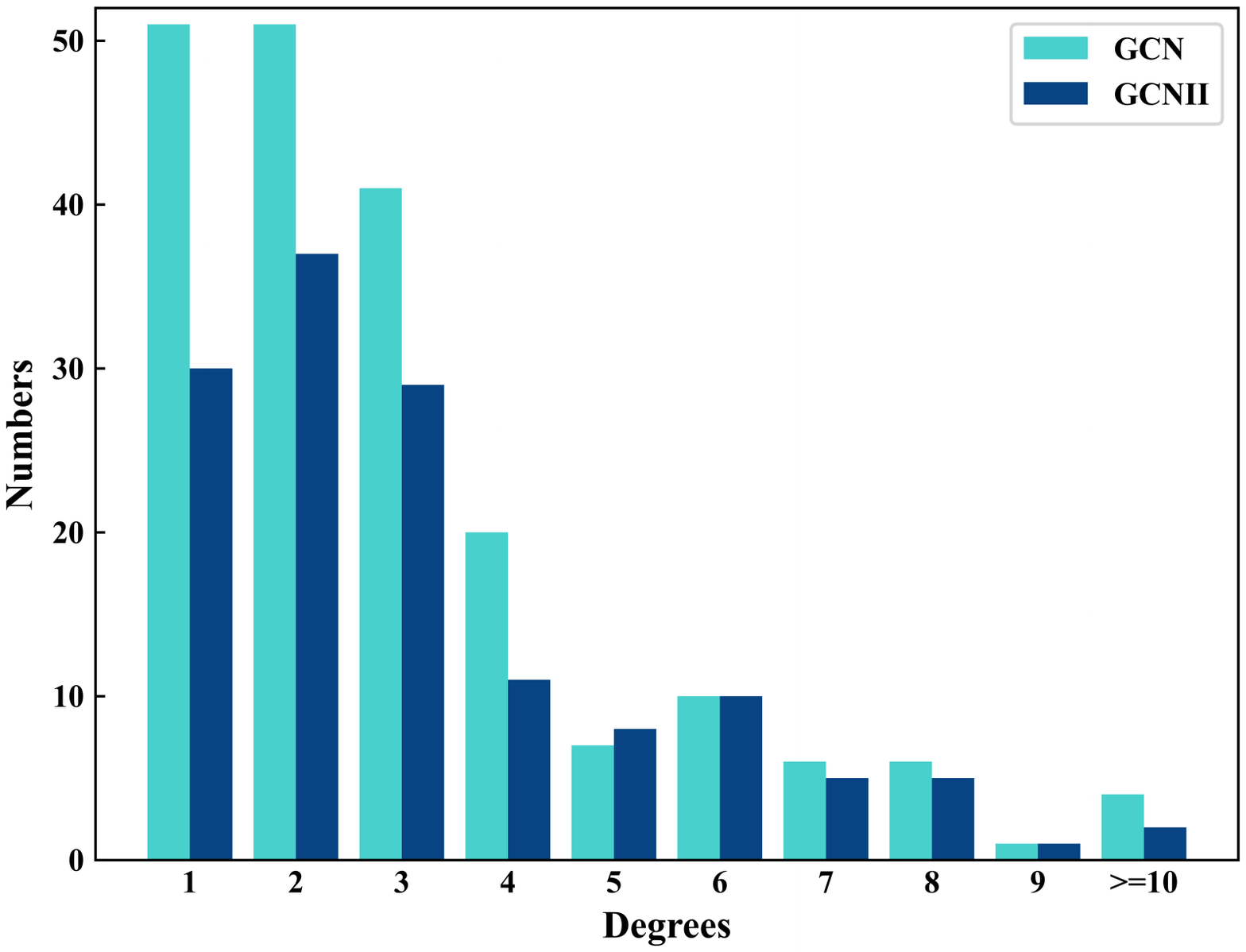}}
    \vspace{-3mm}
    \caption{Degree distribution of misclassified nodes of 2-layer GCN and 64-layer GCNII on Cora.}
    \label{dist}
    \end{center}
    \vspace{-7mm}
\end{figure}

Out-of-distribution generalization refers to the model's ability to maintain strong performance when tested on data that differs from the distribution of the training data.
For graph data, isolated point pairs that are connected only to each other and isolated subgraphs that are connected only internally can be considered out-of-distribution data.
From \cref{dist}, we can observe that the number of misclassified nodes with small degrees of GCNII is significantly reduced.
Many of these nodes are the out-of-distribution data we mentioned above, and there is no path connection between them and the training nodes. 
No matter how deep GCN model is used, the feature information of these nodes cannot be observed during the training process.
Therefore, it demonstrates that GCNII has stronger out-of-distribution generalization ability.

% \section{LLMs for GCNIII.}
% \label{sec:llm}

\end{document}